\documentclass[10pt]{article}
\usepackage{hmi-preprint}
\usepackage{url}
\usepackage{booktabs}
\usepackage{multirow}
\usepackage{tabularx}
\usepackage{colortbl}
\usepackage{pifont}
\usepackage{placeins}
\usepackage{microtype}
\usepackage{marvosym}

\definecolor{cvprblue}{rgb}{0.21,0.49,0.74}
\hypersetup{colorlinks=true,linkcolor=cvprblue,citecolor=cvprblue,urlcolor=cvprblue,
  pdftitle={PINT: A Unified Model with Shared Predictive Intent for Robot Manipulation},
  pdfauthor={Xinyu Guo, Chao Jing, Li Yi, Xingyu Chen, Zhengxing Wu}}

\usepackage{amsmath,amsfonts,bm}

\def\eqref#1{equation~\ref{#1}}

\def\1{\bm{1}}

\DeclareMathAlphabet{\mathsfit}{\encodingdefault}{\sfdefault}{m}{sl}
\SetMathAlphabet{\mathsfit}{bold}{\encodingdefault}{\sfdefault}{bx}{n}

\preprintlogos{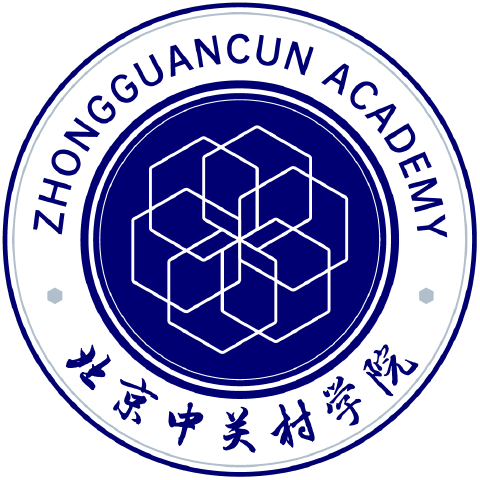}{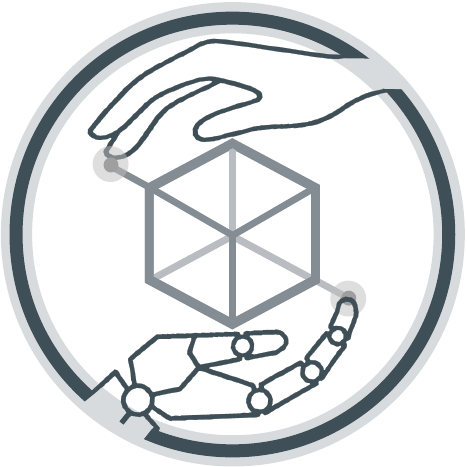}
\preprintdate{October 2026}
\newcommand{\pinttablewidth}{\dimexpr\linewidth-2pc\relax}
\runningtitle{PINT: A Unified Model with Shared Predictive Intent for Robot Manipulation}
\preprintfooter{\textsuperscript{*}Equal contribution.\quad
  \textsuperscript{\Letter}Corresponding authors.\quad
  \textsuperscript{\ensuremath{\ddagger}}Project lead.}
\preprintlinkset
  {https://zgca-hmi-lab.github.io/PINT}
  {https://github.com/ZGCA-HMI-Lab/PINT}
  {}

\makeatletter
\renewcommand{\@preprintlinks}{%
  \vspace*{-4pt}%
  \begin{center}%
    \normalsize
    \begin{tabular}{@{}c@{\hspace{0.55em}}l@{}}%
      \@preprintlinkrow{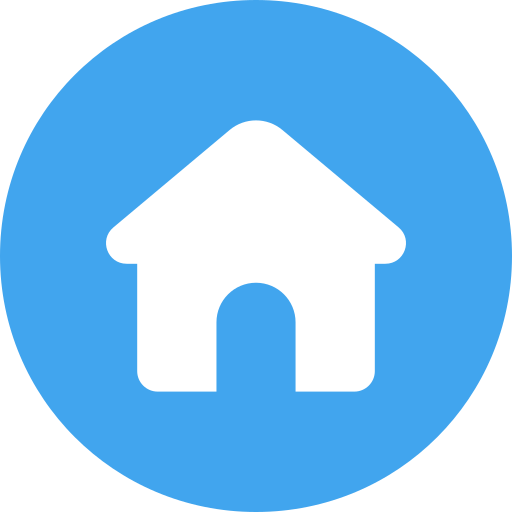}{\@preprintprojecturl}\\[2pt]%
      \@preprintlinkrow{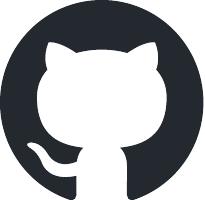}{\@preprintcodeurl}%
    \end{tabular}%
  \end{center}%
  \vspace*{-4pt}%
}
\makeatother

\makeatletter
\renewcommand{\@preprintteaser}{
\par\vspace{8pt}
\begingroup
    \expandafter\def\csname @captype\endcsname{figure}
    \centering
    \includegraphics[
        width=\linewidth,
        trim={135pt 60pt 95pt 95pt},
        clip
    ]{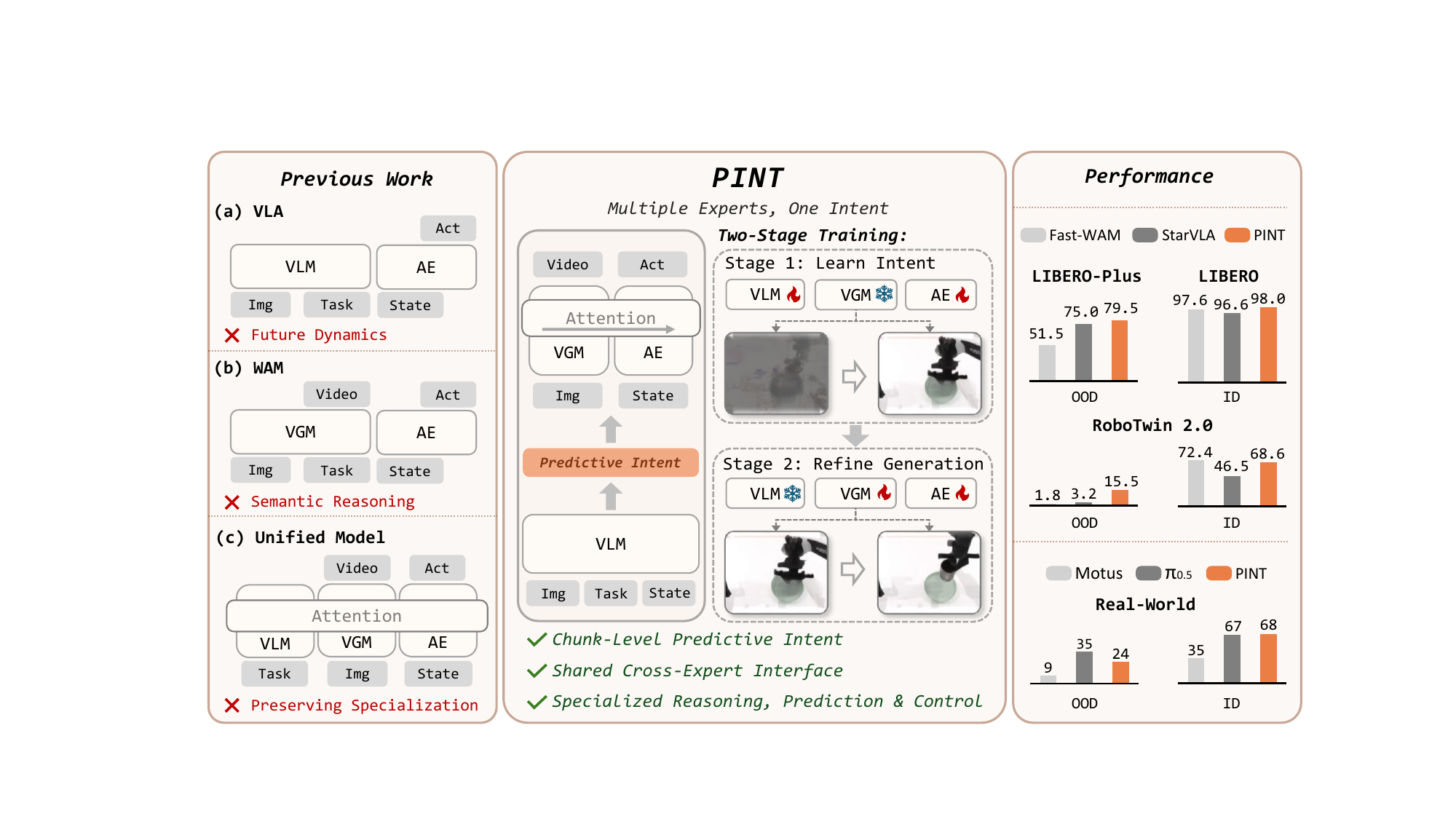}
    \par
\caption{\textbf{Overview of PINT.}
Left: existing approaches to robot manipulation and their limitations.
Middle: PINT coordinates specialized experts through a shared chunk-level Predictive Intent while preserving expert-specific pathways, and learns this Intent with a two-stage training strategy.
Right: selected results across simulation and real-world manipulation under ID/OOD conditions.
For the real-world comparison, Motus~\citep{bi2026motus} and $\pi_{0.5}$~\citep{intelligence2025pi_} use embodied pretraining, whereas PINT does not.}
    \label{fig:concept}
\endgroup
}
\makeatother

\title{PINT: A Unified Model with Shared Predictive Intent for Robot Manipulation}
\author{
Xinyu Guo\textsuperscript{1,2,3,*} \hspace{0.12in}
Chao Jing\textsuperscript{4,3,*} \hspace{0.12in}
Li Yi\textsuperscript{5} \\[3pt]
Xingyu Chen\textsuperscript{3,\Letter,\ensuremath{\ddagger}} \hspace{0.12in}
Zhengxing Wu\textsuperscript{1,\Letter}
}
\affil{
\textsuperscript{1}Institute of Automation, Chinese Academy of Sciences \\
\textsuperscript{2}University of Chinese Academy of Sciences \\
\textsuperscript{3}Zhongguancun Academy \quad
\textsuperscript{4}Fudan University \quad
\textsuperscript{5}Tsinghua University
}

\begin{document}
\maketitle
\begin{abstract}
Generalist robot manipulation can benefit from pretrained vision-language models (VLMs) for semantic reasoning and video generation models (VGMs) for world modeling.
However, existing unified models that directly fuse them through shared attention at multiple layers can make it harder to draw on their specialized capabilities.
We introduce PINT, a unified model that coordinates semantic reasoning, world modeling, and action generation through a shared chunk-level Predictive Intent while preserving expert-specific pathways.
PINT inserts learnable Intent Tokens into the VLM sequence to form this representation, conditioning both the VGM and an Action Expert (AE).
It replaces only the VGM's prompt conditioning and retains the VGM at inference, where the AE reads its representations through one-way attention.
Rather than prescribing what Predictive Intent should encode, PINT learns it jointly from the native prediction objectives of the VLM, VGM, and AE, which provide complementary views of the same upcoming task execution.
Training uses two stages: Stage~1 freezes the VGM so that future video prediction shapes Predictive Intent with task-relevant future dynamics, while Stage~2 fixes the learned Intent and adapts the VGM and AE.
Without embodied pretraining, PINT achieves strong in-distribution (ID) performance and substantially improves out-of-distribution (OOD) generalization, reaching 15.5\% on RoboTwin 2.0-Hard compared with 1.8\% for Fast-WAM and 3.2\% for StarVLA, and 79.5\% on LIBERO-Plus compared with 51.5\% and 75.0\%, respectively.
Across five real-world bimanual tasks, PINT reaches 68\% ID and 24\% OOD success, compared with 35\%/9\% for pretrained Motus and 67\%/35\% for pretrained $\pi_{0.5}$.
\end{abstract}
\section{Introduction}
\label{sec:introduction}

Generalist robot manipulation requires understanding and interacting with complex physical environments across diverse tasks.
Leveraging the semantic knowledge and reasoning capabilities of pretrained vision-language models (VLMs), Vision-Language-Action (VLA) models translate visual and linguistic understanding into robot actions~\citep{brohan2023rt,kim2024openvla,black2024pi_0}.
Yet standard VLAs do not explicitly model future physical dynamics.
In parallel, large-scale video pretraining equips video generation models (VGMs) with strong priors over physical dynamics~\citep{wan2025wan}.
Recent World-Action Models (WAMs)~\citep{ye2026world,yuan2026fast} leverage these priors for future video prediction and action generation, but typically do not exploit the semantic reasoning of VLMs.
Together, these advances motivate unified models that integrate semantic reasoning, world modeling, and action generation.
This raises a central challenge: how to effectively combine the complementary capabilities of pretrained VLMs and VGMs.

Recent unified models often directly fuse vision-language, video, and action representations through shared attention at multiple layers~\citep{bi2026motus,team2026motubrain}.
Yet pretrained VLMs and VGMs acquire their capabilities through different architectures, conditioning schemes, and learning objectives.
These differences lead to specialized internal representations.
Direct fusion forces these representations to serve both expert-specific computation and cross-expert coordination, which can make it harder to draw on each pretrained expert's specialized capabilities.
Several methods instead preserve selected pretrained components or introduce structured interaction pathways~\citep{ma2026internvla,tang2026world}.
These approaches highlight a core tension between expert integration and specialization.
The key question is what to share for coordination and what to preserve for specialization.

To this end, we propose \textbf{PINT} (Fig.~\ref{fig:concept}), a unified model that coordinates semantic reasoning, world modeling, and action generation through a shared Predictive Intent while preserving expert-specific pathways.
Predictive Intent is a compact, chunk-level representation of the upcoming task execution.
PINT inserts learnable Intent Tokens into the VLM sequence to form this representation, which conditions both the VGM and the Action Expert (AE).
PINT replaces only the VGM's prompt conditioning, preserving its pretrained generative pathway, and retains the VGM at inference, where the AE reads its representations through one-way attention.
Rather than prescribing what Predictive Intent should encode, PINT learns it jointly from the native prediction objectives of the VLM, VGM, and AE, which provide complementary views of the same upcoming task execution.
To facilitate learning of Predictive Intent, PINT adopts a two-stage training strategy: Stage~1 freezes the VGM, forcing future video optimization to rely on the Predictive Intent and shaping it with task-relevant future dynamics.
Stage~2 then freezes the VLM and Intent Tokens while adapting the VGM and AE to refine video and action generation.

We compare PINT with representative VLA, WAM, and unified models, including Motus~\citep{bi2026motus} as a representative of direct fusion, across simulation and real-world settings.
Without embodied pretraining, PINT achieves strong in-distribution (ID) performance and substantially improves out-of-distribution (OOD) generalization.
On RoboTwin 2.0-Hard~\citep{chen2025robotwin}, PINT reaches 15.5\% success, compared with 1.8\% for Fast-WAM~\citep{yuan2026fast} and 3.2\% for StarVLA~\citep{starvla2026starvla}.
On LIBERO-Plus~\citep{fei2026libero}, PINT reaches 79.5\% success, compared with 51.5\% for Fast-WAM and 75.0\% for StarVLA.
Across five real-world bimanual tasks, PINT reaches 68\% ID and 24\% OOD success, compared with 35\%/9\% for pretrained Motus~\citep{bi2026motus} and 67\%/35\% for pretrained $\pi_{0.5}$~\citep{intelligence2025pi_}.

Our contributions are summarized as follows:
\begin{itemize}

\item We introduce Predictive Intent, a compact chunk-level representation that serves as a shared interface across specialized experts.
It is learned jointly from the native prediction objectives of the VLM, VGM, and AE without prescribing what it should encode.

\item We propose PINT, which realizes Predictive Intent through learnable Intent Tokens that condition both the VGM and AE, and replaces only the VGM's prompt conditioning, preserving its pretrained generative pathway.
The VGM remains active at inference so that the AE can draw on its representations through one-way attention, while a two-stage strategy lets future video supervision shape Predictive Intent with the VGM frozen before adapting the VGM and AE.

\item Across RoboTwin 2.0, LIBERO, LIBERO-Plus, and real-world bimanual manipulation, PINT achieves strong ID performance and substantially improves OOD generalization without embodied pretraining.
It outperforms comparable VLA and WAM baselines under OOD evaluation, matches pretrained $\pi_{0.5}$ on real-world ID tasks, and outperforms pretrained Motus under both ID and OOD evaluation.

\end{itemize}


\section{Related Work}
\label{sec:related_work}

\subsection{Vision-Language-Action and World-Action Models}

VLA models adapt pretrained VLMs for robot action generation.
RT-2 transfers web-scale vision-language knowledge to robotic control through action tokenization~\citep{brohan2023rt}, while OpenVLA scales this paradigm to diverse robot demonstrations~\citep{kim2024openvla}.
Later work improves continuous action generation with $\pi_0$~\citep{black2024pi_0}, action tokenization with FAST~\citep{pertsch2025fast}, and open-world generalization with $\pi_{0.5}$~\citep{intelligence2025pi_}.
Recent extensions further improve downstream adaptation, temporal memory, and contact-rich manipulation~\citep{guo2026priorvla,shi2026memoryvla,zhang2026feeling}.
Several VLA approaches introduce explicit future supervision, including future image goals in CoT-VLA, future prediction in UP-VLA, and world knowledge forecasting in DreamVLA~\citep{zhao2025cotvla,zhang2025upvla,zhang2025dreamvla}.

A complementary line more directly couples future visual prediction with action generation.
GR-1 and GR-2 jointly model future visual observations and robot actions, leveraging large-scale video generative pretraining for manipulation~\citep{wu2024unleashing,cheang2024gr}.
Recent WAMs extend this direction through unified image and action generation in WorldVLA~\citep{cen2506worldvla}, pretrained video diffusion in DreamZero~\citep{ye2026world}, and training-time video modeling without future denoising at inference in Fast-WAM~\citep{yuan2026fast}.

\subsection{Unified Semantic, World, and Action Modeling}

Recent unified policies combine semantic reasoning, world modeling, and action generation through shared or interacting model components.
Representative approaches couple these capabilities through multi-expert interaction, shared video and action representations, or unified generative backbones, including Motus, MotuBrain, LingBot-VA, LingBot-VA 2.0, and Cosmos 3~\citep{bi2026motus,team2026motubrain,li2026causal,zhang2026native,agarwal2026cosmos}.
In these designs, intermediate representations directly mediate information exchange across capabilities.

A complementary line routes future information through compact representations.
World Tokens maps VLM features into tokens that condition both video prediction and action generation~\citep{tang2026world}, while InternVLA-A1.5 learns foresight tokens in an auxiliary expert that shares attention layers with the VLM and replaces the text conditioning of a frozen pretrained VGM~\citep{ma2026internvla}.
Both remove the video branch at inference~\citep{tang2026world,ma2026internvla}.
WLA-0 conditions its world expert with meta-queries~\citep{yang2026world}, DIAL defines intent as latent visual foresight~\citep{chen2026dial}, and LAWA represents future intention through compact latent actions~\citep{li2026latent}.
Other methods represent future evolution through latent queries, visual subgoals, or structured world representations~\citep{luo2026being,chen2026lawam,liu2026oa,li2026egowam}.
Compared with World Tokens and InternVLA-A1.5, PINT forms Predictive Intent directly within the VLM sequence, without an external adapter or an auxiliary expert sharing attention with the VLM~\citep{tang2026world,ma2026internvla}.
PINT first learns Predictive Intent with the VGM frozen, then fixes the learned Intent and adapts the VGM, which remains active at inference and provides representations to the AE through one-way attention.

\section{Method}
\label{sec:method}

\subsection{Problem Formulation}

We consider generalist robot manipulation, where a policy maps multi-view images $\mathbf{I}_t$, a task instruction $\ell$, and proprioceptive state $\mathbf{s}_t$ to a short-horizon action chunk.
PINT further models future dynamics through future video prediction.
Given the same context, a unified model $\mathcal{F}_\theta$ predicts future video and robot actions:
\begin{equation}
    \left(
        \mathbf{v}_{t+1:t+H_v},
        \mathbf{a}_{t:t+H_a-1}
    \right)
    =
    \mathcal{F}_\theta
    \left(
        \mathbf{I}_t,\ell,\mathbf{s}_t
    \right),
    \label{eq:problem_formulation}
\end{equation}
where $H_v$ and $H_a$ denote the video and action horizons, respectively.

During training, the VLM additionally predicts an autoregressive suffix of discrete action tokens, $\mathbf{y}^{\mathrm{AR}}_{t,1:N}$, as auxiliary supervision.
This suffix is omitted at inference.

\subsection{Overview}

\begin{figure}[t]
    \centering
    \includegraphics[
        width=\linewidth,
        trim={20pt 140pt 45pt 140pt},
        clip
    ]{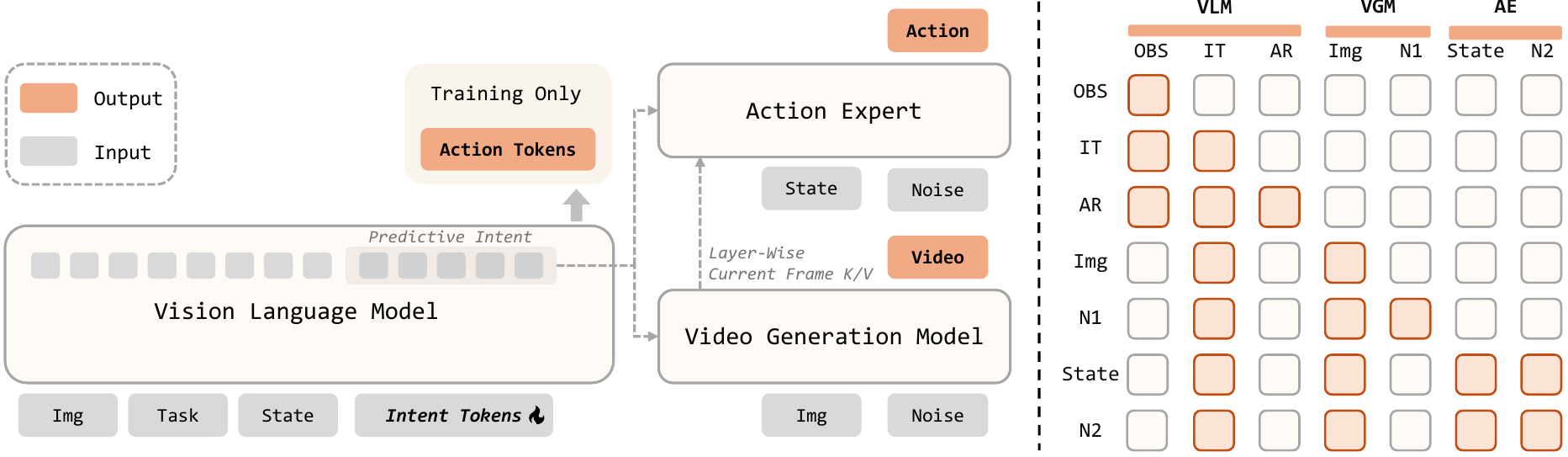}
    \vspace{-5mm}
    \caption{\textbf{PINT architecture and attention design.}
    Left: PINT inserts learnable Intent Tokens into the VLM sequence to form Predictive Intent, which conditions both the VGM and AE while preserving their expert-specific pathways and inputs. The AE receives layer-wise current-frame K/V from the VGM.
    Right: attention patterns among the VLM, VGM, and AE. OBS, IT, and AR denote the observation prefix, Intent Tokens, and training-only autoregressive suffix of action tokens. Img and $N_1$ denote the VGM image and noise inputs, while State and $N_2$ denote the AE state and action-noise inputs.}
    \label{fig:overview}
\end{figure}

As illustrated in Fig.~\ref{fig:overview}, PINT organizes unified prediction around a shared Predictive Intent:
\begin{equation}
    \mathbf{z}_t
    =
    \mathrm{VLM}
    \left(
        \mathbf{I}_t,\ell,\mathbf{s}_t;\mathbf{Q}
    \right),
    \qquad
    \begin{cases}
        \mathbf{v}_{t+1:t+H_v}
        =
        \mathrm{VGM}
        \left(
            \bar{\mathbf{I}}_t;\mathbf{z}_t
        \right), \\[2pt]
        \mathbf{a}_{t:t+H_a-1}
        =
        \mathrm{AE}
        \left(
            \mathbf{s}_t,\mathbf{h}_t;\mathbf{z}_t
        \right)
    \end{cases}.
    \label{eq:pint_overview}
\end{equation}

Here, $\mathbf{Q}$ denotes the learnable Intent Tokens, $\mathbf{z}_t$ the Predictive Intent, and $\mathbf{h}_t$ the layer-wise VGM key/value features from the current frame.
Separate linear projections map Predictive Intent to the conditioning spaces of the VGM and AE.
The Intent Tokens are placed at the end of the multimodal prefix, and their final-layer hidden states form the Predictive Intent.
The VGM combines the current frame with Predictive Intent to predict future video, while the AE combines Predictive Intent with proprioceptive state and VGM features from the current frame to generate continuous actions.
During training, the VLM additionally receives autoregressive supervision through a suffix placed after the Intent Tokens.
PINT is trained in two stages, first learning Predictive Intent with a frozen VGM and then adapting the VGM and AE while keeping the learned Intent fixed.

\subsection{Predictive Intent}
\label{sec:intent}

Let $\mathbf{Q}=\{\mathbf{q}_m\}_{m=1}^{M}$ denote $M$ learnable Intent Tokens placed at the end of the VLM prefix, and let $\mathbf{H}^{L}_t$ denote the final-layer hidden states of the full prefix.
We partition these states according to their token groups:
\begin{equation}
    \mathbf{H}^{L}_t
    =
    \left[
        \mathbf{H}^{L}_{I,t},
        \mathbf{H}^{L}_{\ell,t},
        \mathbf{H}^{L}_{s,t},
        \mathbf{z}_t
    \right],
    \label{eq:predictive_intent}
\end{equation}
where $L$ denotes the final VLM layer, and $\mathbf{H}^{L}_{I,t}$, $\mathbf{H}^{L}_{\ell,t}$, and $\mathbf{H}^{L}_{s,t}$ correspond to the image, instruction, and state tokens, respectively.
The final group $\mathbf{z}_t \in \mathbb{R}^{M \times d}$ corresponds to the Intent Token positions and defines the Predictive Intent, where $d$ is the VLM hidden dimension.

During training, the autoregressive suffix follows the Intent Tokens under the causal attention mask.
Thus, the Predictive Intent $\mathbf{z}_t$ depends only on the multimodal prefix, while subsequent autoregressive tokens can attend to it.
Given the target sequence $\mathbf{y}^{\mathrm{AR}}_{t,1:N}$, the VLM is trained with the standard next-token prediction objective:
\begin{equation}
    \mathcal{L}_{\mathrm{AR}}
    =
    -\sum_{n=1}^{N}
    \log
    p_{\theta}
    \left(
        y^{\mathrm{AR}}_{t,n}
        \mid
        \mathbf{I}_t,\ell,\mathbf{s}_t,\mathbf{Q},
        \mathbf{y}^{\mathrm{AR}}_{t,<n}
    \right).
    \label{eq:ar_loss}
\end{equation}
The autoregressive supervision uses the VLM's native next-token prediction objective over a suffix containing only discrete action tokens.
Together with the video and action objectives introduced below, this supervision helps shape Predictive Intent into a shared representation.
The two-stage schedule separates Intent learning from subsequent expert adaptation, as described in Sec.~\ref{sec:training}.

\subsection{Intent-Conditioned Future Video Prediction}
\label{sec:vgm}

Large-scale video pretraining equips the VGM with strong priors over physical dynamics.
PINT leverages these priors by replacing the VGM's original prompt conditioning with Predictive Intent, which represents the upcoming task execution.
The multi-view images $\mathbf{I}_t$ are composed into a single current frame $\bar{\mathbf{I}}_t$, while $\mathbf{z}_t$ is linearly projected into the VGM conditioning space and supplied through its existing cross-attention layers.
This changes only the conditioning signal while preserving the pretrained VGM's generative mechanism.

Following the native flow matching objective~\citep{lipman2022flow,wan2025wan}, the VGM predicts future video in the latent space of the pretrained video VAE by learning a velocity field.
Let $\mathbf{r}$ denote the latent representation of the target future video.
Given Gaussian noise $\boldsymbol{\epsilon}_v$ and a flow time $\tau \in (0,1)$, we construct
\begin{equation}
    \mathbf{r}_{\tau}
    =
    (1-\tau)\mathbf{r}
    +
    \tau\boldsymbol{\epsilon}_v.
    \label{eq:video_flow}
\end{equation}


The VGM predicts the corresponding velocity field $f_\theta^v$ conditioned on the current frame and Predictive Intent:
\begin{equation}
    \mathcal{L}_{\mathrm{video}}
    =
    \mathbb{E}_{\mathbf{r},\boldsymbol{\epsilon}_v,\tau}
    \left[
        \left\|
            f^{v}_{\theta}
            \left(
                \mathbf{r}_{\tau},
                \tau,
                \bar{\mathbf{I}}_t,
                \mathbf{z}_t
            \right)
            -
            \left(
                \boldsymbol{\epsilon}_v-\mathbf{r}
            \right)
        \right\|_2^2
    \right].
    \label{eq:video_loss}
\end{equation}

During Stage~1, the VGM remains frozen, while $\mathcal{L}_{\mathrm{video}}$ backpropagates through it to the Predictive Intent.

In the default PINT configuration, future video denoising is omitted at inference.
The VGM performs a single forward pass on the current frame conditioned on Predictive Intent, and the resulting layer-wise K/V features are cached for all AE denoising steps.

\subsection{Intent-Conditioned Action Generation}
\label{sec:action}

The AE is conditioned on the shared Predictive Intent, aligning action generation with future video prediction around the same upcoming execution.
It also receives proprioceptive state $\mathbf{s}_t$ and attends to layer-wise VGM key/value features from the current frame, providing fine-grained visual information for control.
This interaction is one-way in the forward pass: the AE attends to VGM representations, while VGM tokens never attend to AE tokens.
In Stage~2, gradients from the action objective also propagate through these features to the VGM.

Continuous action generation follows the same flow matching formulation.
Given Gaussian noise $\boldsymbol{\epsilon}_a$ and a flow time $\tau \in (0,1)$, we construct the noisy action chunk
\begin{equation}
    \mathbf{a}_{\tau}
    =
    (1-\tau)\mathbf{a}_{t:t+H_a-1}
    +
    \tau\boldsymbol{\epsilon}_a,
    \label{eq:action_flow}
\end{equation}
and train the AE to predict the corresponding velocity field $f_\theta^a$:
\begin{equation}
    \mathcal{L}_{\mathrm{action}}
    =
    \mathbb{E}_{\mathbf{a}_{t:t+H_a-1},\boldsymbol{\epsilon}_a,\tau}
    \left[
        \left\|
            f_\theta^a
            \left(
                \mathbf{a}_{\tau},
                \tau,
                \mathbf{s}_t,
                \mathbf{h}_t,
                \mathbf{z}_t
            \right)
            -
            \left(
                \boldsymbol{\epsilon}_a
                -
                \mathbf{a}_{t:t+H_a-1}
            \right)
        \right\|_2^2
    \right].
    \label{eq:action_loss}
\end{equation}

During Stage~1, the action objective provides continuous control supervision to Predictive Intent, complementing the autoregressive and future video objectives.

\subsection{Two-Stage Training}
\label{sec:training}

\begin{figure}[t]
    \centering
    \includegraphics[
        width=0.85\linewidth,
        trim={95pt 73pt 158pt 90pt},
        clip
    ]{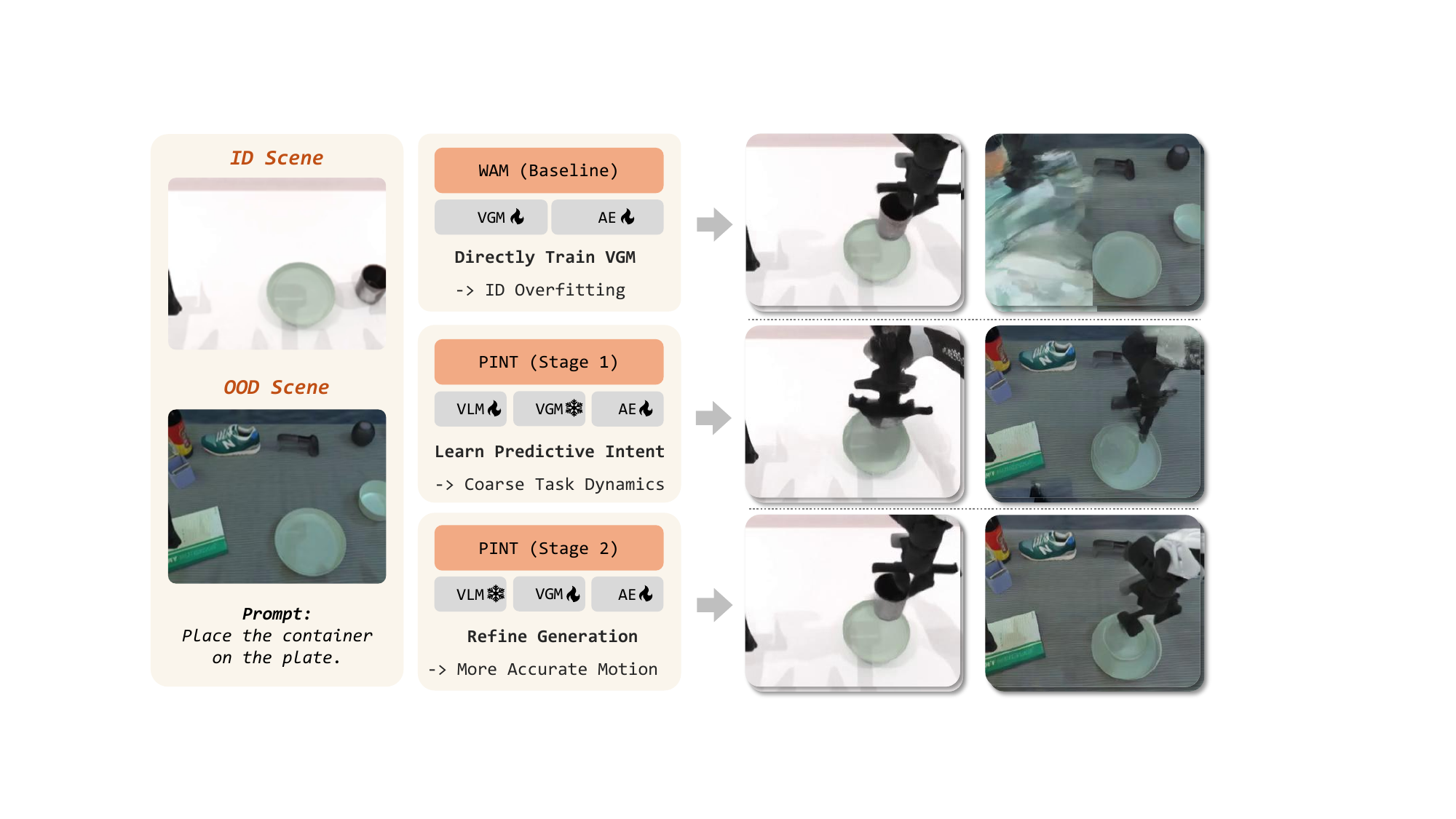}\par
    \vspace{-8pt}
\caption{\textbf{Future Video Prediction across ID/OOD Scenes and Training Stages.}
We visualize predictions from the VGM trained in Fast-WAM~\citep{yuan2026fast} and PINT after Stage~1 and Stage~2 on the same ID and OOD inputs.
With the VGM frozen and no direct text conditioning, instruction semantics reach the VGM only through Predictive Intent, already yielding coarse task dynamics.
Stage~2 fixes the learned Intent and adapts the VGM for more accurate motion.}
    \label{fig:qualitative}
    \vspace{-16pt}
\end{figure}

We optimize the three prediction objectives in two stages:
\begin{equation}
\begin{aligned}
    \mathcal{L}_{\mathrm{stage1}}
    &=
    \lambda_{\mathrm{AR}}\mathcal{L}_{\mathrm{AR}}
    +
    \lambda_{\mathrm{video}}\mathcal{L}_{\mathrm{video}}
    +
    \lambda_{\mathrm{action}}\mathcal{L}_{\mathrm{action}},
    \\[3pt]
    \mathcal{L}_{\mathrm{stage2}}
    &=
    \lambda_{\mathrm{video}}\mathcal{L}_{\mathrm{video}}
    +
    \lambda_{\mathrm{action}}\mathcal{L}_{\mathrm{action}} .
\end{aligned}
\label{eq:two_stage_objective}
\end{equation}
Here, $\lambda_{\mathrm{AR}}$, $\lambda_{\mathrm{video}}$, and $\lambda_{\mathrm{action}}$ balance the corresponding objectives.

\textbf{Stage~1: Learning Predictive Intent.}
The pretrained VGM is frozen while the VLM, Intent Tokens, and AE are optimized with $\mathcal{L}_{\mathrm{stage1}}$.
With the VGM fixed, optimization of the future video objective must act through the Intent-conditioning pathway, allowing its gradients to shape Predictive Intent with task-relevant future dynamics.
Together with autoregressive and action supervision, this yields a shared predictive representation of the upcoming task execution.

\textbf{Stage~2: Refining Generation.}
The VLM and Intent Tokens are then frozen, fixing the learned mapping from multimodal context to Predictive Intent.
The VGM and AE are optimized with $\mathcal{L}_{\mathrm{stage2}}$ to refine future video and action generation conditioned on the fixed Predictive Intent.
Fig.~\ref{fig:qualitative} compares future video prediction from the VGM used in Fast-WAM with PINT across training stages.

We additionally evaluate PINT-VA, a variant that keeps Stage~1 unchanged.
In Stage~2, the AE attends one-way to VGM features from all video latents rather than only the current frame, and at inference the future video and action latents are denoised synchronously.

\begingroup
\makeatletter
\renewcommand{\section}{\@startsection{section}{1}{\z@}
  {12pt plus 2pt minus 2pt}{10pt plus 2pt minus 2pt}{\large\bfseries}}
\renewcommand{\subsection}{\@startsection{subsection}{2}{\z@}
  {9pt plus 2pt minus 2pt}{9pt plus 2pt minus 2pt}{\elvbf}}
\renewcommand{\paragraph}{\@startsection{paragraph}{4}{\z@}
  {8pt plus 2pt minus 1pt}{-1em}{\normalfont\normalsize\bfseries}}
\makeatother
\section{Experiments}
\label{sec:experiments}

\suppressfloats[t]


\begin{table}[!t]
\centering
\setlength{\belowcaptionskip}{\baselineskip}
\caption{\textbf{Performance on LIBERO and LIBERO-Plus.}
Success rates (\%) on LIBERO (ID) and LIBERO-Plus (OOD).
Best and second-best results in each group are bolded and underlined.}
\label{tab:libero_plus}
\label{tab:libero}

\small
\setlength{\tabcolsep}{2.5pt}

\resizebox{\pinttablewidth}{!}{
\begin{tabular}{l|ccccc|cccccccc}
\toprule
\multirow{2}{*}{\textbf{Method}}
& \multicolumn{5}{c|}{\textbf{LIBERO}}
& \multicolumn{8}{c}{\textbf{LIBERO-Plus}} \\
\cmidrule(lr){2-6}\cmidrule(lr){7-14}
& \textbf{Spatial} & \textbf{Object} & \textbf{Goal} & \textbf{Long} & \textbf{Avg.}
& \textbf{Cam.}
& \textbf{Robot}
& \textbf{Lang.}
& \textbf{Light}
& \textbf{BG}
& \textbf{Noise}
& \textbf{Layout}
& \textbf{Total} \\
\midrule

\multicolumn{14}{l}{\footnotesize\itshape With Embodied Pretraining} \\

$\pi_{0}$~\citep{black2024pi_0}
& 96.8 & 98.8 & 95.8 & 85.2 & 94.2
& 13.8
& 6.0
& 58.8
& 85.0
& 81.4
& 79.0
& 68.9
& 53.6 \\

$\pi_{0}$-FAST~\citep{pertsch2025fast}
& 96.4 & 96.8 & 88.6 & 60.2 & 85.5
& 65.1
& 21.6
& 61.0
& 73.2
& 73.2
& 74.4
& 68.8
& 61.6 \\

RIPT-VLA~\citep{tan2025interactive}
& 98.6 & 98.6 & \textbf{99.0} & 93.8 & 97.5
& 55.2
& 31.2
& 77.6
& 88.4
& 91.6
& 73.5
& 74.2
& 68.4 \\

OpenVLA-OFT~\citep{kim2025fine}
& 97.6 & 98.4 & 97.9 & 94.5 & 97.1
& 56.4
& 31.9
& 79.5
& 88.7
& \underline{93.3}
& 75.8
& 74.2
& 69.6 \\

MemoryVLA++~\citep{shi2026memoryvla++}
& \textbf{99.8} & \textbf{100.0} & \underline{98.2} & 96.0 & \textbf{98.5}
& 36.4
& 68.9
& \underline{88.7}
& 93.8
& 90.6
& 63.5
& 83.8
& 73.1 \\

Motus~\citep{bi2026motus}
& 96.8 & \underline{99.8} & 96.6 & \textbf{97.6} & \underline{97.7}
& 48.0
& \textbf{88.0}
& \textbf{89.0}
& 83.0
& 78.0
& 57.0
& \textbf{89.0}
& 76.0 \\

TBD-VLA~\citep{lee2026tbd}
& 97.6 & 99.6 & 97.4 & \underline{96.6} & \underline{97.7}
& \textbf{87.8}
& 60.4
& 77.4
& \underline{95.8}
& 88.8
& \textbf{89.9}
& 84.4
& \underline{83.5} \\

$\pi_{0.5}$~\citep{intelligence2025pi_}
& \underline{98.8} & 98.2 & 98.0 & 92.4 & 96.9
& \underline{75.4}
& \underline{77.5}
& 85.6
& \textbf{96.9}
& \textbf{94.6}
& \underline{89.7}
& \underline{85.7}
& \textbf{85.7} \\

\midrule

\multicolumn{14}{l}{\footnotesize\itshape Without Embodied Pretraining} \\

Fast-WAM~\citep{yuan2026fast}
& \textbf{98.2} & \textbf{100.0} & 97.0 & 95.2 & 97.6
& 16.4
& 44.5
& 68.9
& 78.2
& 53.7
& 37.7
& 60.7
& 51.5 \\

VLA-Adapter~\citep{wang2026vla}
& 97.8 & 99.2 & 97.2 & 95.0 & 97.3
& 36.2
& 37.9
& 74.6
& 70.6
& 76.1
& 58.0
& 69.7
& 59.1 \\

TBD-VLA (Scratch)~\citep{lee2026tbd}
& --- & --- & --- & --- & ---
& 29.4
& 62.9
& 52.1
& 89.4
& 88.8
& 61.7
& 79.0
& 66.2 \\


StarVLA-OFT~\citep{starvla2026starvla}
& 97.8 & 98.6 & 96.2 & 93.8 & 96.6
& 47.0
& 60.1
& 87.0
& \textbf{96.3}
& \textbf{95.3}
& 73.1
& 79.2
& 75.0 \\

ResVLA~\citep{zhong2026noise}
& 96.8 & 98.6 & \underline{97.4} & 92.4 & 96.3
& 49.8
& 59.9
& \textbf{88.5}
& 90.5
& \underline{94.9}
& 76.8
& 79.0
& 75.3 \\

\rowcolor{gray!15}
\textbf{PINT}
& \underline{98.0} & \underline{99.6} & \underline{97.4} & \textbf{97.0} & \underline{98.0}
& \underline{59.2}
& \underline{73.7}
& 84.4
& \underline{94.0}
& 91.4
& \underline{80.8}
& \textbf{81.2}
& \underline{79.5} \\

\rowcolor{gray!15}
\textbf{PINT-VA}
& \textbf{98.2} & \underline{99.6} & \textbf{98.0} & \underline{96.8} & \textbf{98.2}
& \textbf{61.4}
& \textbf{76.1}
& \underline{87.1}
& 93.9
& 93.3
& \textbf{82.3}
& \underline{81.0}
& \textbf{81.0} \\

\bottomrule
\end{tabular}
}

\vspace{-10pt}
\end{table}

\subsection{Experimental Setup}
\label{sec:setup}

\paragraph{Implementation details.}

We use Qwen3.5-4B~\citep{qwen2026qwen35} and Wan2.2-5B~\citep{wan2025wan} as the VLM and VGM backbones.
PINT conditions the AE on layer-wise current-frame VGM features, while PINT-VA extends conditioning to all video latents and synchronously denoises future video and action latents at inference.
On a single NVIDIA A100 80GB GPU (BF16, batch size 1), end-to-end generation of a 32-step action chunk with 10 denoising steps takes 266, 331, and 776 ms for Fast-WAM, PINT, and PINT-VA, respectively.
PINT therefore incurs only moderate additional latency over Fast-WAM, while PINT-VA trades substantially higher inference cost for stronger performance.
Additional implementation, evaluation, and latency details are provided in
Appendices~\ref{app:implementation}, \ref{app:evaluation}, and~\ref{app:inference_efficiency}.
\par\endgroup

\suppressfloats[t]


\begin{table}[!t]

\centering

\setlength{\belowcaptionskip}{\baselineskip}
\caption{\textbf{Performance on RoboTwin 2.0.}
Success rates (\%) under Easy and Hard, with all methods trained on clean demonstrations.
PT denotes embodied pretraining.
Best and second-best in each group are bolded and underlined.
$\dagger$ denotes our reproduction.}

\label{tab:robotwin_clean2random}
\label{tab:robotwin}

\normalsize

\renewcommand{\arraystretch}{1.08}

\begin{tabularx}{0.78\linewidth}{
>{\hsize=2.2\hsize\linewidth=\hsize\raggedright\arraybackslash}X
>{\centering\arraybackslash}c
>{\hsize=0.6\hsize\linewidth=\hsize\centering\arraybackslash}X
>{\hsize=0.6\hsize\linewidth=\hsize\centering\arraybackslash}X
>{\hsize=0.6\hsize\linewidth=\hsize\centering\arraybackslash}X
}

\toprule

\textbf{Method}
& \textbf{PT}
& \textbf{Easy}
& \textbf{Hard}
& \textbf{Avg.} \\

\midrule

\multicolumn{5}{@{}l}{\small\itshape With Embodied Pretraining} \\[-1pt]

RDT~\citep{liu2024rdt}
& $\checkmark$
& 34.5
& 13.7
& 24.1 \\

$\pi_0$~\citep{black2024pi_0}
& $\checkmark$
& 46.4
& 16.3
& 31.4 \\

GalaxeaVLA~\citep{jiang2025galaxea}
& $\checkmark$
& 62.7
& 12.7
& 37.7 \\

ABot-M0~\citep{yang2026abot}
& $\checkmark$
& 57.4
& \underline{30.4}
& 43.9 \\

X-VLA~\citep{zheng2025xvla}
& $\checkmark$
& 68.0
& 20.9
& 44.5 \\

X-WAM~\citep{guo2026xwam}
& $\checkmark$
& \underline{70.0}
& 25.8
& \underline{47.9} \\

$\pi_{0.5}$~\citep{intelligence2025pi_}
& $\checkmark$
& \textbf{70.7}
& \textbf{46.0}
& \textbf{58.4} \\

\midrule

\multicolumn{5}{@{}l}{\small\itshape Without Embodied Pretraining} \\[-1pt]

StarVLA~\citep{starvla2026starvla}
& $\times$
& 46.5
& 3.2
& 24.9 \\

AHA-WAM~\citep{cai2026ahawam}
& $\times$
& 64.3
& 3.2
& 33.8 \\

Fast-WAM$^\dagger$~\citep{yuan2026fast}
& $\times$
& \textbf{72.4}
& 1.8
& 37.1 \\

Evo-1~\citep{lin2025evo}
& $\times$
& 65.7
& 12.5
& 39.1 \\

\rowcolor{gray!10}
\textbf{PINT}
& $\times$
& 68.6
& \underline{15.5}
& \underline{42.1} \\

\rowcolor{gray!10}
\textbf{PINT-VA}
& $\times$
& \underline{71.9}
& \textbf{18.1}
& \textbf{45.0} \\

\bottomrule

\end{tabularx}

\vspace{-6pt}
\end{table}

\paragraph{Benchmarks and evaluation.}

We evaluate PINT on LIBERO~\citep{liu2023libero}, LIBERO-Plus~\citep{fei2026libero}, RoboTwin 2.0~\citep{chen2025robotwin}, and five real-world manipulation tasks.
For LIBERO, we jointly train on the four standard suites, evaluate on LIBERO for ID performance, and directly transfer the same policy to LIBERO-Plus for OOD evaluation.
For RoboTwin 2.0, we train on all 50 tasks using the official clean demonstrations and evaluate Easy and Hard as ID and OOD settings, respectively.
For real-world experiments, we evaluate ID and OOD conditions, with OOD jointly varying lighting, background, and object location.

\paragraph{Baselines.}

We compare PINT with representative VLA, WAM, and unified-model baselines.
Embodied pretraining provides additional robot experience and can improve downstream performance and generalization, particularly under distribution shift~\citep{intelligence2025pi_,yang2026abot}.
We therefore group methods by embodied pretraining.
Methods without embodied pretraining serve as our primary comparison, while embodied-pretrained methods provide strong references.
We highlight Fast-WAM~\citep{yuan2026fast}, StarVLA~\citep{starvla2026starvla}, and Motus~\citep{bi2026motus} as representative baselines for WAMs, VLAs, and unified models with direct fusion, respectively.
Fast-WAM uses the same Wan2.2-5B VGM as PINT, StarVLA uses a Qwen-family VLM, and Motus uses both a Qwen-family VLM and the same VGM, together with embodied pretraining (result provenance in Appendix~\ref{app:provenance}).

\subsection{Main Results}
\label{sec:main_results}

\paragraph{LIBERO and LIBERO-Plus.}

As shown in Table~\ref{tab:libero}, PINT achieves 98.0\% average success on LIBERO and 79.5\% on LIBERO-Plus without embodied pretraining.
On LIBERO-Plus, PINT substantially outperforms Fast-WAM (51.5\%) and surpasses StarVLA-OFT (75.0\%) and ResVLA (75.3\%), the strongest baseline without embodied pretraining.
It also exceeds several embodied-pretrained methods, including Motus (76.0\%) and MemoryVLA++ (73.1\%).
PINT-VA further improves the averages to 98.2\% on LIBERO and 81.0\% on LIBERO-Plus.

\paragraph{RoboTwin 2.0.}

As shown in Table~\ref{tab:robotwin}, PINT achieves 68.6\% success on Easy and 15.5\% on Hard without embodied pretraining.
Compared with Fast-WAM, PINT is slightly lower on Easy (68.6\% vs.\ 72.4\%) but achieves substantially higher Hard success (15.5\% vs.\ 1.8\%), resulting in a higher average success rate (42.1\% vs.\ 37.1\%).
PINT also surpasses Evo-1 (12.5\%), the strongest baseline without embodied pretraining on Hard, and exceeds its 39.1\% average success rate.
PINT-VA further improves Easy, Hard, and average success to 71.9\%, 18.1\%, and 45.0\%, respectively (repeated evaluations and per-task results in Appendices~\ref{app:repeated_evaluation} and~\ref{app:robotwin_per_task}).

\begin{table}[t]
    \centering
    \setlength{\belowcaptionskip}{\baselineskip}
\caption{\textbf{Real-world robot manipulation results.}
Success rates (\%) under ID/OOD evaluation.
$\pi_{0.5}$~\citep{intelligence2025pi_} and Motus~\citep{bi2026motus} use embodied pretraining, while PINT does not.
Best and second-best are bolded and underlined.}
    \label{tab:real_world}

    \small
    \setlength{\tabcolsep}{4pt}
    \renewcommand{\arraystretch}{1.15}
    \begin{tabularx}{\pinttablewidth}{@{}cc|*{5}{>{\centering\arraybackslash}X}|>{\centering\arraybackslash}X}
        \toprule
        \multirow{2}{*}{Eval.} & \multirow{2}{*}{Method}
        & Turn On & Stack & Place & Store & Fold & Avg. \\
        & & Light & Cups & Pens & Block & Towel & Success \\
        \midrule
        & $\pi_{0.5}$ & \textbf{90} & \underline{85} & \textbf{65} & \underline{75} & \textbf{20} & \underline{67} \\
        ID & Motus & 50 & 60 & 30 & 35 & 0 & 35 \\
        & \cellcolor{gray!15}\textbf{PINT}
        & \cellcolor{gray!15}\textbf{90} & \cellcolor{gray!15}\textbf{90}
        & \cellcolor{gray!15}\underline{60} & \cellcolor{gray!15}\textbf{85}
        & \cellcolor{gray!15}\underline{15} & \cellcolor{gray!15}\textbf{68} \\
        \midrule
        & $\pi_{0.5}$ & \textbf{30} & \textbf{70} & \textbf{20} & \textbf{45} & \textbf{10} & \textbf{35} \\
        OOD & Motus & 5 & 20 & 5 & 15 & 0 & 9 \\
        & \cellcolor{gray!15}\textbf{PINT}
        & \cellcolor{gray!15}\underline{20} & \cellcolor{gray!15}\underline{55}
        & \cellcolor{gray!15}\underline{10} & \cellcolor{gray!15}\underline{30}
        & \cellcolor{gray!15}\underline{5} & \cellcolor{gray!15}\underline{24} \\
        \bottomrule
    \end{tabularx}

    \vspace{10pt}
\end{table}

\begin{table}[!t]
    \centering
\caption{\textbf{Ablation studies of PINT.}
All variants are trained on the 50-task clean demonstrations from RoboTwin 2.0 and evaluated under the Easy and Hard settings.
We report success rates (\%).}
    \label{tab:ablation}

    \begin{minipage}{\pinttablewidth}
    \small
    \setlength{\tabcolsep}{4pt}
    \renewcommand{\arraystretch}{1.00}

    \begin{minipage}[t]{0.48\linewidth}
        \vspace{0pt}
        \centering
        \parbox[t][1.0\baselineskip][t]{\linewidth}{\centering\textbf{(a) Intent Design}}
        \par\vspace{0pt}
        \renewcommand{\arraystretch}{1.25}
        \begin{tabularx}{\linewidth}{>{\raggedright\arraybackslash}X c c}
            \toprule
            \textbf{Setting} & \textbf{Easy} & \textbf{Hard} \\
            \midrule
            Prompt-conditioned VGM & 62.4 & 11.1 \\
            Split Intent & 65.1 & 11.8 \\
            \rowcolor{gray!15}
            \textbf{PINT} & \textbf{68.6} & \textbf{15.5} \\
            \bottomrule
        \end{tabularx}
    \end{minipage}\hfill
    \begin{minipage}[t]{0.48\linewidth}
        \vspace{0pt}
        \centering
        \parbox[t][1.0\baselineskip][t]{\linewidth}{\centering\textbf{(b) Supervision and Expert Interaction}}
        \par\vspace{0pt}
        \begin{tabularx}{\linewidth}{>{\raggedright\arraybackslash}X c c}
            \toprule
            \textbf{Setting} & \textbf{Easy} & \textbf{Hard} \\
            \midrule
            w/o Stage~1 Video Loss & 63.2 & 11.5 \\
            w/o VGM-to-AE Features & 65.9 & 14.6 \\
            w/o AR Loss & 67.3 & 12.2 \\
            \rowcolor{gray!15}
            \textbf{PINT} & \textbf{68.6} & \textbf{15.5} \\
            \bottomrule
        \end{tabularx}
    \end{minipage}

    \par\vspace{4pt}

    \begin{minipage}[t]{0.48\linewidth}
        \vspace{0pt}
        \centering
        \parbox[t][1.0\baselineskip][t]{\linewidth}{\centering\textbf{(c) Two-Stage Training}}
        \par\vspace{0pt}
        \begin{tabularx}{\linewidth}{>{\raggedright\arraybackslash}X c c}
            \toprule
            \textbf{Setting} & \textbf{Easy} & \textbf{Hard} \\
            \midrule
            Stage~1 Only & 58.5 & 13.1 \\
            Single-Stage Training & 65.5 & 14.4 \\
            \rowcolor{gray!15}
            \textbf{PINT} & \textbf{68.6} & \textbf{15.5} \\
            \bottomrule
        \end{tabularx}
    \end{minipage}\hfill
    \begin{minipage}[t]{0.48\linewidth}
        \vspace{0pt}
        \centering
        \parbox[t][1.0\baselineskip][t]{\linewidth}{\centering\textbf{(d) Intent Capacity}}
        \par\vspace{0pt}
        \begin{tabularx}{\linewidth}{>{\raggedright\arraybackslash}X c c}
            \toprule
            \textbf{Setting} & \textbf{Easy} & \textbf{Hard} \\
            \midrule
            16 Intent Tokens & 67.9 & 15.0 \\
            64 Intent Tokens & 67.1 & 14.7 \\
            \rowcolor{gray!15}
            \textbf{PINT (32 Tokens)} & \textbf{68.6} & \textbf{15.5} \\
            \bottomrule
        \end{tabularx}
    \end{minipage}
    \end{minipage}
\end{table}

\paragraph{Real-world experiments.}

Table~\ref{tab:real_world} summarizes five real-world tasks (setup and rollouts in Appendices~\ref{app:realworld} and~\ref{app:realworld_qualitative}).
Without embodied pretraining, PINT achieves 68\% ID success, matching pretrained $\pi_{0.5}$ (67\%) and substantially outperforming pretrained Motus (35\%).
Under OOD evaluation, PINT reaches 24\%, outperforming Motus (9\%) while remaining below $\pi_{0.5}$ (35\%).

\subsection{Ablation Studies}
\label{sec:ablation}

Table~\ref{tab:ablation} evaluates the main design choices of PINT on RoboTwin 2.0 Easy and Hard (variant details in Appendix~\ref{app:training_variants}).
Prompt-conditioned VGM uses the original prompt instead of Predictive Intent, reducing success from 68.6\%/15.5\% to 62.4\%/11.1\%.
On LIBERO/LIBERO-Plus, success drops from 98.0\%/79.5\% to 95.9\%/63.4\%, with a substantially larger OOD drop (Appendix~\ref{app:libero_ablation}).
Fig.~\ref{fig:qualitative} shows that Fast-WAM produces plausible ID predictions but degrades under OOD shift, whereas PINT after Stage~1 captures coarse task-directed dynamics and Stage~2 further refines the motion, with closed-loop ID/OOD trajectories and open-loop predictions across five tasks in Appendix~\ref{app:future_video}.
Split Intent, with separate VGM- and AE-specific Intent groups, reaches 65.1\%/11.8\%.

Removing Stage~1 future-video supervision reduces success to 63.2\%/11.5\% on RoboTwin and
97.1\%/75.2\% on LIBERO/LIBERO-Plus, while removing autoregressive supervision yields
67.3\%/12.2\%.
Removing VGM-to-AE features yields 65.9\%/14.6\%, indicating that direct access to VGM representations at inference provides information beyond Predictive Intent.
Stage~1 Only and Single-Stage Training achieve 58.5\%/13.1\% and 65.5\%/14.4\%, respectively.

\subsection{Predictive Intent Visualization}
\label{sec:intent_visualization}

\begin{figure}[!t]
    \centering
    \includegraphics[
        width=0.65\linewidth,
        trim={330pt 225pt 308pt 125pt},
        clip
    ]{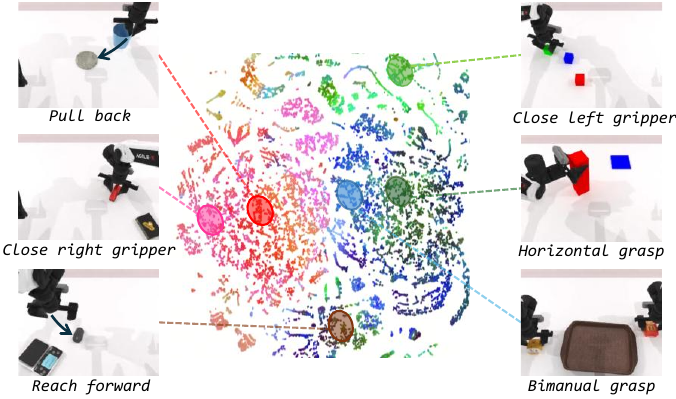}\par
    \vspace{-8pt}
    \caption{\textbf{Predictive Intent visualization.}
    We project ground-truth action chunks into a two-dimensional t-SNE space and color each point using a three-dimensional PCA projection of its corresponding Predictive Intent.
    Representative behaviors from highlighted regions are shown.}
    \label{fig:intent_visualization}
    \vspace{-8pt}
\end{figure}

Fig.~\ref{fig:intent_visualization} projects ground-truth action chunks with t-SNE~\citep{vandermaaten2008visualizing} and independently colors each point using a three-dimensional PCA projection of the mean-pooled Intent Tokens, as detailed in Appendix~\ref{app:intent_visualization}.
Representative examples from highlighted regions exhibit distinct manipulation behaviors, while the structured color variation suggests that Predictive Intent captures behavioral structure across diverse trajectories.

\FloatBarrier

\section{Conclusion}
\label{sec:conclusion}

We introduced PINT, which coordinates semantic reasoning, world modeling, and action generation through a shared chunk-level Predictive Intent while preserving expert-specific pathways.
PINT learns this Intent from the native prediction objectives of the VLM, VGM, and AE through a two-stage strategy that first shapes it with a frozen pretrained VGM and then adapts the VGM and AE.
Across simulation and real-world manipulation, PINT achieves strong ID performance and improved OOD generalization without embodied pretraining.
These results suggest that a compact shared predictive interface can coordinate specialized pretrained models for robot manipulation.

\paragraph{Limitations.}
Our evaluation focuses on robot manipulation and a single pairing of VLM and VGM backbones.
Combining PINT with embodied pretraining and extending it to broader embodied settings and larger-scale training remain future work.

\bibliographystyle{unsrtnat}
\bibliography{references}

\clearpage
\appendix
\setlength{\belowcaptionskip}{6pt}
\section{Implementation Details}
\label{app:implementation}
\label{app:implementation_notes}
\label{app:training_details}

\subsection{Model Configuration and Inference}
\label{app:model_config}

\paragraph{Model components.}
We instantiate the VLM with Qwen3.5-4B and the VGM with Wan2.2-TI2V-5B, including its pretrained video DiT and VAE.
The AE is a state-conditioned action DiT based on the ActionDiT architecture of Fast-WAM, with 30 Transformer layers and a hidden dimension of 1024.
It operates on continuous action chunks and predicts the corresponding flow velocity.
Across all benchmarks, PINT uses $M=32$ Intent Tokens with a VLM hidden dimension of 2560.
Two independent linear projections with bias map the same Predictive Intent to the conditioning spaces of the VGM and AE:
\[
P_{\mathrm{VGM}}:\mathbb{R}^{2560}\rightarrow\mathbb{R}^{3072},
\qquad
P_{\mathrm{AE}}:\mathbb{R}^{2560}\rightarrow\mathbb{R}^{1024}.
\]
No additional MLP or normalization layer is applied in either projection.

\paragraph{Input organization.}
The VLM receives each camera view as an independent image, together with the task instruction and the current robot state discretized as text.
The 32 Intent Tokens are appended to the multimodal prefix, followed by the training-only autoregressive action suffix.
At inference, the autoregressive suffix is omitted.
PINT therefore extracts Predictive Intent with a single VLM prefix forward pass and does not autoregressively generate action tokens or intermediate reasoning traces.

The VGM uses a different visual organization from the VLM.
Rather than processing the camera views independently, it receives a composed multi-view frame.
Table~\ref{tab:input_configuration} summarizes the benchmark-specific visual layouts and state/action dimensions.
All VLM views are independently letterboxed to $224\times224$ before being passed to the VLM.
For RoboTwin 2.0 and the real-world ALOHA setup, the head-camera image is resized to $256\times320$, while each wrist-camera image is resized to $128\times160$.
The head view is placed above the two wrist views, producing a $384\times320$ VGM input.
For LIBERO, the $224\times224$ main and wrist views are placed side by side, producing a $224\times448$ VGM input.

\begin{table}[!htbp]
\centering
\small
\caption{\textbf{Benchmark-specific input configuration.}
Image sizes are reported as height $\times$ width.
The VLM receives camera views independently, whereas the VGM operates on a composed multi-view frame.}
\label{tab:input_configuration}
\begin{tabularx}{\pinttablewidth}{
@{}
>{\raggedright\arraybackslash}p{0.15\linewidth}
>{\raggedright\arraybackslash}p{0.19\linewidth}
>{\raggedright\arraybackslash}X
>{\centering\arraybackslash}p{0.11\linewidth}
>{\centering\arraybackslash}p{0.11\linewidth}
@{}}
\toprule
Benchmark & Camera views & VGM composition & State dim. & Action dim. \\
\midrule
RoboTwin 2.0
& Head + left wrist + right wrist
& Head above two wrist views, $384\times320$
& 14 & 14 \\

LIBERO
& Main + wrist
& Main and wrist views side by side, $224\times448$
& 8 & 7 \\

Real-world ALOHA
& Head + left wrist + right wrist
& Same layout as RoboTwin 2.0, $384\times320$
& 14 & 14 \\
\bottomrule
\end{tabularx}
\end{table}

\paragraph{Video and action configuration.}
For future video prediction, all benchmarks use nine frames sampled at
\[
t, t+4, t+8, \ldots, t+32,
\]
comprising the current frame and eight future frames over 32 original data steps.
We report this horizon in data steps rather than seconds because the underlying control frequency differs across datasets.
The AE predicts a continuous action chunk of $H_a=32$ steps for all benchmarks.
The current proprioceptive state is encoded by a two-layer MLP with a SiLU activation into a single 1024-dimensional state token, which is prepended to the 32 action tokens during AE computation and removed before action prediction.

The AE receives VGM representations through layer-wise one-way attention.
Both the AE and VGM contain 30 Transformer layers.
At AE layer $\ell$, the action queries attend to the K/V representations produced by VGM layer $\ell$, in addition to the AE's own K/V representations.
Thus, the AE accesses VGM features throughout the network rather than only from the final VGM layer.

\paragraph{Inference.}
Default PINT first performs a VLM forward pass to obtain Predictive Intent.
The composed current frame is then encoded by the pretrained video VAE, and the VGM performs a single forward pass conditioned on Predictive Intent.
The resulting layer-wise current-frame K/V representations are cached and reused throughout action generation.
The AE generates the 32-step action chunk using 10 explicit-Euler denoising steps with the shifted flow-time schedule used in our implementation ($\mathrm{shift}=5$), after which the predicted actions are mapped back to the original action space.
Default PINT does not denoise future-video latents at inference.

PINT-VA uses the same model components and keeps Stage~1 unchanged.
During Stage~2 and inference, the AE additionally accesses layer-wise VGM features from all video latents.
The nine-frame video sequence is temporally compressed by the Wan VAE into three latent time positions, consisting of one current-frame position and two future positions, each with its own spatial latent grid.
The current-frame latent remains fixed, while future-video and action latents are synchronously updated using the same 10-step shifted flow-time schedule.
Future-video latents are not decoded back to RGB during standard action inference.

\subsection{Optimization and Hyperparameters}
\label{app:hyperparameters}

\paragraph{Optimization.}
We use AdamW for all experiments with $\beta_1=0.9$, $\beta_2=0.95$, $\epsilon=10^{-8}$, and weight decay $0.01$.
Weight decay is applied to matrix-valued weights, while bias, normalization, and other non-matrix parameters are excluded.
Each training stage independently uses a linear warm-up over the first 5\% of optimization followed by cosine learning-rate decay.
Gradients are clipped to a maximum norm of $1.0$.
Training uses BF16 precision with DeepSpeed ZeRO-1 on 8 NVIDIA A100 80GB GPUs, without gradient accumulation.

We use the same module-wise learning-rate schedule across RoboTwin 2.0, LIBERO, and the real-world ALOHA experiments.
Table~\ref{tab:module_lr} summarizes the peak and final learning rates for each trainable component.
During Stage~1, the VGM DiT and video VAE are frozen, while the VLM, Intent Tokens, AE, and both Intent projectors are optimized.
During Stage~2, the VLM, Intent Tokens, and video VAE are frozen, while the VGM DiT, AE, and both Intent projectors are optimized.

\begin{table}[!htbp]
\centering
\small
\caption{\textbf{Module-wise learning rates.}
Each entry reports the peak learning rate $\rightarrow$ final learning rate after cosine decay.}
\label{tab:module_lr}
\begin{tabular}{lcc}
\toprule
Module & Stage~1 & Stage~2 \\
\midrule
VLM + Intent Tokens
& $10^{-5}\rightarrow10^{-6}$
& Frozen \\

Intent $\rightarrow$ VGM projector
& $10^{-5}\rightarrow10^{-6}$
& $10^{-4}\rightarrow10^{-6}$ \\

Intent $\rightarrow$ AE projector
& $10^{-5}\rightarrow10^{-6}$
& $10^{-4}\rightarrow10^{-6}$ \\

AE + state encoder
& $10^{-5}\rightarrow10^{-6}$
& $10^{-4}\rightarrow10^{-6}$ \\

VGM DiT
& Frozen
& $10^{-4}\rightarrow10^{-6}$ \\

Video VAE
& Frozen
& Frozen \\
\bottomrule
\end{tabular}
\end{table}

\paragraph{Training objectives.}
We use the same loss weights across all benchmarks.
During Stage~1, the autoregressive, video, and action objectives are weighted by
$\lambda_{\mathrm{AR}}=0.1$,
$\lambda_{\mathrm{video}}=0.1$, and
$\lambda_{\mathrm{action}}=1.0$, respectively.
Stage~2 does not execute the autoregressive branch and optimizes only the video and action objectives with
$\lambda_{\mathrm{video}}=0.1$ and
$\lambda_{\mathrm{action}}=1.0$.

The autoregressive loss is averaged over valid supervised tokens.
The action and video objectives are averaged over valid action elements and future-video latent elements, respectively, and use the timestep-dependent weighting of the flow scheduler.
The video objective is applied only to future-video latents, excluding the current-frame latent, and padded elements are masked from the corresponding losses.

\paragraph{Training budget.}
Table~\ref{tab:training_budget} summarizes the training budget for each benchmark.
Stage~1 uses a per-GPU batch size of 16, corresponding to a global batch size of 128 across 8 GPUs.
Stage~2 uses a per-GPU batch size of 12 and a global batch size of 96.
RoboTwin 2.0 and the real-world ALOHA experiments are trained for 5 epochs in each stage, while LIBERO is trained for 10 epochs per stage.

\begin{table}[!htbp]
\centering
\small
\caption{\textbf{Training budget across benchmarks.}
Global batch sizes correspond to 8 GPUs with no gradient accumulation.}
\label{tab:training_budget}
\begin{tabular}{lccc}
\toprule
Hyperparameter & RoboTwin 2.0 & LIBERO & Real-World ALOHA \\
\midrule
Stage~1 epochs & 5 & 10 & 5 \\
Stage~2 epochs & 5 & 10 & 5 \\
Stage~1 batch size / GPU & 16 & 16 & 16 \\
Stage~2 batch size / GPU & 12 & 12 & 12 \\
Stage~1 global batch size & 128 & 128 & 128 \\
Stage~2 global batch size & 96 & 96 & 96 \\
\bottomrule
\end{tabular}
\end{table}

\paragraph{Checkpoint selection.}
Stage~2 is initialized from the final Stage~1 checkpoint.
For all reported results, we evaluate the final Stage~2 checkpoint rather than selecting checkpoints according to validation or test performance.
PINT-VA follows the same optimization configuration and training budget as PINT.

\subsection{Two-Stage Training and Variant Details}
\label{app:training_variants}

\paragraph{Initialization and stage transition.}
Stage~1 initializes the VLM from Qwen3.5-4B and the VGM and video VAE from Wan2.2-TI2V-5B.
Following Fast-WAM, the AE backbone is initialized by copying or interpolating weights from the pretrained Wan DiT, while newly introduced components, including the action encoder, state encoder, and output head, are initialized separately.
The two Intent projectors are randomly initialized.
The 32 Intent Token embeddings are initialized with the mean of the original Qwen vocabulary embeddings and are subsequently optimized as independent learnable tokens.

During Stage~1, the VGM DiT and video VAE are frozen, while the VLM, Intent Tokens, AE, and both Intent projectors are optimized jointly.
Stage~2 is initialized from the final Stage~1 checkpoint without reinitializing the AE or projectors.
The VLM, Intent Tokens, and video VAE are then frozen, while the VGM DiT, AE, and both Intent projectors remain trainable.
The autoregressive branch is not executed during Stage~2.
As described in Appendix~\ref{app:hyperparameters}, the final Stage~2 checkpoint is used for evaluation.

\paragraph{Prompt-conditioned VGM.}
This variant replaces only the conditioning input to the VGM.
Instead of Predictive Intent, the VGM receives the task instruction through the original Wan T5 text-conditioning pathway.
The AE continues to use Predictive Intent together with proprioceptive state, and the 32 Intent Tokens remain otherwise unchanged.
The Intent-to-VGM projector is bypassed for VGM conditioning.
All other components follow the same two-stage schedule, training budget, and checkpoint-selection protocol as PINT.

\paragraph{Split Intent.}
This variant replaces the shared Intent group with two expert-specific groups containing 32 VGM Intent Tokens and 32 AE Intent Tokens:
\[
\text{multimodal prefix}
\rightarrow
Q_{\mathrm{VGM}}
\rightarrow
Q_{\mathrm{AE}}
\rightarrow
\text{AR suffix}.
\]
The VGM and AE read their corresponding token groups through separate projectors.
Under the causal VLM mask, the AE-specific Intent Tokens can attend to the preceding VGM-specific group, while the reverse interaction is not permitted.
The remaining expert connections and training procedure are unchanged.

\paragraph{Loss ablations.}
For \textit{w/o Stage~1 Video Loss}, we remove the future-video prediction objective only during Stage~1.
Stage~2 restores the standard video and action objectives.
This variant therefore isolates the role of video supervision in shaping Predictive Intent during Stage~1 rather than removing video modeling throughout training.
For \textit{w/o AR Loss}, the autoregressive branch and its objective are removed during Stage~1, while the video and continuous-action objectives remain unchanged.
Stage~2 is identical to the default PINT configuration.

\paragraph{Expert-interaction ablation.}
For \textit{w/o VGM-to-AE Features}, the AE no longer attends to the layer-wise K/V representations of the VGM.
It remains conditioned on Predictive Intent and proprioceptive state, while the VGM continues to optimize the future-video objective.
The two-stage training schedule is otherwise unchanged.

\paragraph{Training-schedule ablations.}
\textit{Stage~1 Only} directly evaluates the final Stage~1 checkpoint without performing Stage~2.
Inference remains identical to default PINT: the VLM first produces Predictive Intent, the VGM performs a single current-frame forward pass to produce cached layer-wise K/V features, and the AE generates the action chunk with 10 denoising steps.

\textit{Single-Stage Training} jointly optimizes the VLM, Intent Tokens, VGM DiT, AE, and both Intent projectors from initialization, while keeping the video VAE frozen.
It uses the autoregressive, video, and action objectives throughout training.
On RoboTwin 2.0, it is trained for 10 epochs, matching the total $5+5$ epoch budget of the default two-stage schedule.
The VLM and Intent Tokens use a learning rate of $10^{-5}\rightarrow10^{-6}$, while the VGM, AE, and both Intent projectors use $10^{-4}\rightarrow10^{-6}$.
Its inference procedure remains the default current-frame PINT configuration.

\paragraph{Intent capacity.}
The 16- and 64-token variants change only the size of the shared Intent group from the default $M=32$.
Both the VGM and AE continue to condition on the full shared group.
The projector dimensions remain $2560\rightarrow3072$ for the VGM and $2560\rightarrow1024$ for the AE.
All other architectural choices, objectives, optimization settings, and training budgets are kept unchanged.

\FloatBarrier
\section{Benchmarks, Data, and Evaluation}
\label{app:evaluation}
\label{app:evaluation_protocol}
\label{app:training_data}
\label{app:task_details}

\begin{figure}[t]
    \centering
    \includegraphics[
        width=\linewidth,
        trim={85pt 155pt 125pt 10pt},
        clip
    ]{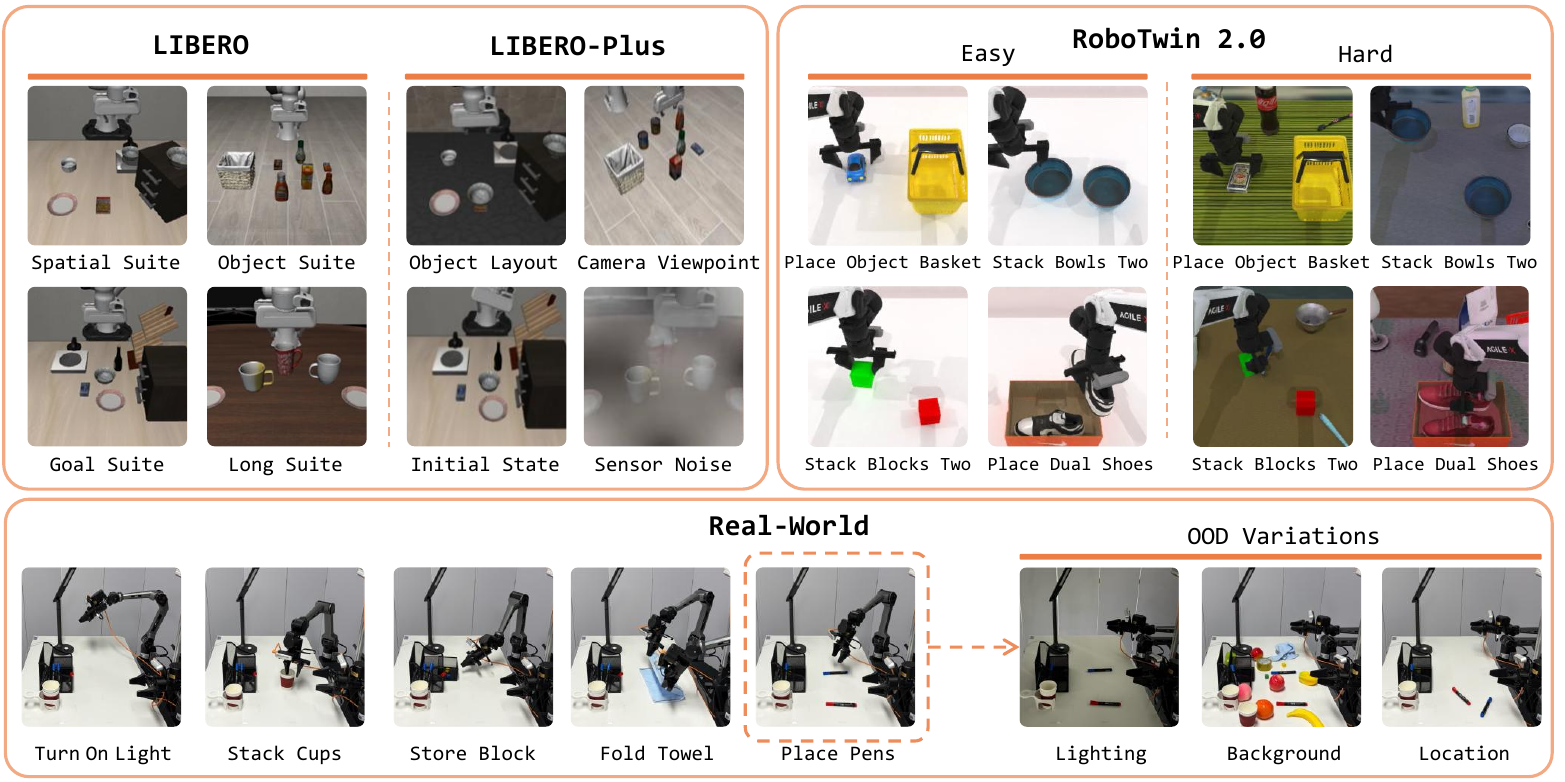}
    \vspace{-15pt}
    \caption{\textbf{Experimental overview.}
    PINT is evaluated on LIBERO~\citep{liu2023libero}, LIBERO-Plus~\citep{fei2026libero}, RoboTwin 2.0~\citep{chen2025robotwin}, and five real-world bimanual tasks under ID/OOD settings.}
    \label{fig:experiments}
    \vspace{0pt}
\end{figure}

Fig.~\ref{fig:experiments} summarizes the benchmarks and real-world evaluation settings.
\subsection{RoboTwin 2.0}
\label{app:robotwin}
\label{app:robotwin_evaluation}
\label{app:robotwin_tasks}

RoboTwin 2.0~\citep{chen2025robotwin} is a bimanual manipulation benchmark built around the Aloha-AgileX dual-arm robot.
We jointly train a single PINT model on all 50 tasks using 50 official clean demonstrations per task, resulting in 2,500 demonstrations.
The tasks cover object interaction, tool use, stacking, handover, lifting, and coordinated dual-arm manipulation.

We evaluate the final checkpoint on all 50 tasks under both Easy and Hard.
Each evaluation uses 100 rollouts per task and setting with the official simulator success signal.
The same checkpoint trained on clean demonstrations is directly evaluated under both settings without additional adaptation.
Easy corresponds to the clean ID environment, whereas Hard follows the clean-to-randomized protocol and introduces domain randomization in background appearance, table clutter, lighting, and table height.
Table~\ref{tab:app_robotwin2_easy_hard_config} summarizes the configuration differences.

\begin{table}[!ht]
\centering
\small
\setlength{\tabcolsep}{6pt}
\caption{
\textbf{RoboTwin 2.0 Easy and Hard evaluation configurations.}
The Easy setting uses the clean environment, while the Hard setting applies the official domain randomization.
}
\label{tab:app_robotwin2_easy_hard_config}

\begin{tabular}{lcc}
\toprule
Field & Easy / ID & Hard / OOD \\
\midrule
\multicolumn{3}{l}{\textit{Domain randomization}} \\
Background randomization & Off & On \\
Table clutter & Off & On \\
Clean background rate & 1.00 & 0.02 \\
Table-height perturbation & 0 & 0.03 \\
Lighting randomization & Off & On \\
Extreme lighting rate & 0 & 0.02 \\
\midrule
\multicolumn{3}{l}{\textit{Shared environment setup}} \\
Robot embodiment & Aloha-AgileX & Aloha-AgileX \\
Head / wrist camera & D435 / D435 & D435 / D435 \\
RGB observation & Yes & Yes \\
End-effector state / qpos & Yes / Yes & Yes / Yes \\
\bottomrule
\end{tabular}
\end{table}
\subsection{LIBERO and LIBERO-Plus}
\label{app:libero}
\label{app:libero_evaluation}
\label{app:libero_tasks}

For LIBERO~\citep{liu2023libero}, we jointly train a single model on Spatial, Object, Goal, and Long using 50 demonstrations per task.
These suites respectively evaluate spatial reasoning, object-centric manipulation, goal-conditioned behaviors, and long-horizon task execution.
We evaluate the final checkpoint on all four suites, conducting 50 rollouts per task with the official initial states, reset procedure, and success signal.

We directly evaluate the same LIBERO-trained checkpoint on LIBERO-Plus~\citep{fei2026libero} without additional training or adaptation.
Following the official protocol, each perturbed task instance is evaluated once across seven shift categories: camera viewpoint, robot initial state, language instruction, lighting condition, background texture, sensor noise, and object layout.
Success is determined using the official success signal.

\paragraph{Aggregate metrics.}
For standard LIBERO, Table~\ref{tab:libero} reports the average success rate across the four suites.
For PINT and PINT-VA, the LIBERO-Plus Total score is computed over all evaluation trials rather than as the arithmetic mean of the seven category-level success rates.
PINT succeeds in 7,974 of 10,030 trials (79.5\%), while PINT-VA succeeds in 8,129 of 10,030 trials (81.0\%).
LIBERO-Plus aggregate scores for external baselines follow their original reported sources.
\subsection{Real-World ALOHA Tasks}
\label{app:realworld}
\label{app:realworld_evaluation}
\label{app:realworld_tasks}

We evaluate PINT on five real-world bimanual manipulation tasks using an ALOHA platform.
The task suite covers switch operation, object stacking, object placement, articulated-object manipulation, and deformable-object manipulation.
The training set contains 700 demonstrations, with per-task counts summarized in Table~\ref{tab:realworld_training_data}.
Table~\ref{tab:realworld_tasks} lists the corresponding language instructions.

We evaluate the final checkpoint under both ID and OOD conditions, with 20 rollouts per task and condition.
Before each rollout, the scene is manually reset according to the corresponding evaluation setting.
OOD evaluation jointly varies lighting, background appearance, and initial object locations while preserving the task semantics and robot embodiment.
Task success is determined using the criteria in Table~\ref{tab:realworld_success_criteria}.

\begin{table}[!t]
\centering
\small
\setlength{\tabcolsep}{12pt}
\caption{
\textbf{Real-world training data for PINT.}
Number of demonstrations collected for each ALOHA manipulation task.
}
\label{tab:realworld_training_data}

\begin{tabular}{lc}
\toprule
Task & Demonstrations \\
\midrule
Turn On Light & 50 \\
Stack Cups & 50 \\
Place Pens & 150 \\
Store Block & 150 \\
Fold Towel & 300 \\
\midrule
Total & 700 \\
\bottomrule
\end{tabular}

\par\vspace{\baselineskip}
\normalsize
\setlength{\tabcolsep}{10pt}
\caption{\textbf{Real-world ALOHA tasks and language instructions.}}
\label{tab:realworld_tasks}

\small
\begin{tabular}{ll}
\toprule
Task & Language Instruction \\
\midrule
Turn On Light & Turn on the light. \\
Stack Cups & Stack the cups. \\
Place Pens & Place the blue and red pens in the pen holder. \\
Store Block & Put the block in the drawer and close it. \\
Fold Towel & Fold the towel. \\
\bottomrule
\end{tabular}

\par\vspace{\baselineskip}
\normalsize
\setlength{\tabcolsep}{8pt}
\caption{
\textbf{Success criteria for the real-world ALOHA tasks.}
Each rollout is considered successful when the corresponding task-specific criterion is satisfied.
}
\label{tab:realworld_success_criteria}

\small
\begin{tabular}{lp{0.68\linewidth}}
\toprule
Task & Success Criterion \\
\midrule
Turn On Light &
The light is turned on. \\

Stack Cups &
The cups are stacked and remain stable after manipulation. \\

Place Pens &
Both the blue and red pens are placed inside the pen holder. \\

Store Block &
The target block is placed inside the drawer and the drawer is fully closed. \\

Fold Towel &
The towel is folded into the target compact rectangular shape without major misalignment or unfolding. \\
\bottomrule
\end{tabular}
\end{table}

\FloatBarrier

\section{Baselines and Result Provenance}
\label{app:provenance}

\subsection{Main-Table Result Sources}
\label{app:main_table_sources}

We document the public sources used for the external baseline results in
Tables~\ref{tab:libero} and~\ref{tab:robotwin}.
When a method's original paper does not report the corresponding benchmark, we use a public external evaluation that matches the reported variant and evaluation setting.
Our own evaluations are identified separately in Appendix~\ref{app:reproductions}.

\begin{table}[!htbp]
\centering
\caption{\textbf{Result provenance for Table~\ref{tab:libero}.}
Each entry identifies the source of the corresponding LIBERO or LIBERO-Plus result.}
\label{tab:libero_provenance}

\footnotesize
\begin{tabularx}{\pinttablewidth}{
@{}
>{\raggedright\arraybackslash}p{0.20\linewidth}
>{\raggedright\arraybackslash}X
>{\raggedright\arraybackslash}X
@{}}
\toprule
Method & LIBERO source & LIBERO-Plus source \\
\midrule

$\pi_0$
&
OpenVLA-OFT~\citep{kim2025fine}
&
Official LIBERO-Plus leaderboard
\\

$\pi_0$-FAST
&
OpenVLA-OFT~\citep{kim2025fine}
&
Official LIBERO-Plus leaderboard
\\

RIPT-VLA
&
RIPT-VLA~\citep{tan2025interactive}
&
Official LIBERO-Plus leaderboard
\\

OpenVLA-OFT
&
OpenVLA-OFT~\citep{kim2025fine}, Table~I
&
Official LIBERO-Plus leaderboard
\\

MemoryVLA++
&
MemoryVLA++~\citep{shi2026memoryvla++}, Table~I
&
MemoryVLA++~\citep{shi2026memoryvla++}, Table~V
\\

Motus
&
MoWAM~\citep{wang2026mowam}, Table~I
&
MoWAM~\citep{wang2026mowam}, Table~III
\\

TBD-VLA
&
TBD-VLA~\citep{lee2026tbd}
&
TBD-VLA~\citep{lee2026tbd}
\\

$\pi_{0.5}$
&
Official OpenPI LIBERO result
&
WAM--VLA~\citep{zhang2026worldaction}, Table~4
\\

\midrule

Fast-WAM
&
Fast-WAM~\citep{yuan2026fast}
&
WAM--VLA~\citep{zhang2026worldaction}, Table~4
\\

VLA-Adapter
&
VLA-Adapter~\citep{wang2026vla}
&
UniMPA~\citep{li2026unimpa}, Table~II
\\

TBD-VLA (Scratch)
&
--
&
TBD-VLA~\citep{lee2026tbd}
\\

StarVLA-OFT
&
Official StarVLA LIBERO benchmark documentation (Qwen3-VL-OFT)
&
3DThinkVLA~\citep{shi20263dthinkvla} (Qwen3-VL-OFT)
\\

ResVLA
&
ResVLA~\citep{zhong2026noise}
&
ResVLA~\citep{zhong2026noise}
\\

PINT
&
\multicolumn{2}{l}{Our evaluation}
\\

PINT-VA
&
\multicolumn{2}{l}{Our evaluation}
\\

\bottomrule
\end{tabularx}
\end{table}

For MemoryVLA++, the original work reports Spatial, Object, Goal, Long-10, and Long-90 separately.
To match the standard four-suite LIBERO protocol used in Table~\ref{tab:libero}, we compute the average over Spatial, Object, Goal, and Long-10, excluding Long-90, which gives 98.5\%.
The LIBERO-Plus value of 73.1\% is taken directly from the zero-shot evaluation in the original paper.
For Motus, both the standard LIBERO and LIBERO-Plus results are external evaluations reported by MoWAM~\citep{wang2026mowam} rather than our reproduction.
For StarVLA-OFT, the LIBERO row corresponds to the Qwen3-VL-OFT variant in the official StarVLA documentation, and the LIBERO-Plus row uses the matched Qwen3-VL-OFT evaluation reported by 3DThinkVLA~\citep{shi20263dthinkvla}.

\begin{table}[!htbp]
\centering
\footnotesize
\caption{\textbf{Result provenance for Table~\ref{tab:robotwin}.}
Leaderboard values refer to the official RoboTwin 2.0 leaderboard snapshot used for this submission.}
\label{tab:robotwin_provenance}
\begin{tabularx}{\pinttablewidth}{
@{}
>{\raggedright\arraybackslash}p{0.22\linewidth}
>{\raggedright\arraybackslash}X
>{\raggedright\arraybackslash}p{0.29\linewidth}
@{}}
\toprule
Method & Source & Note \\
\midrule

RDT
&
Official RoboTwin 2.0 leaderboard
&
Clean-to-clean / clean-to-randomized
\\

$\pi_0$
&
Official RoboTwin 2.0 leaderboard
&
Clean-to-clean / clean-to-randomized
\\

GalaxeaVLA
&
Official RoboTwin 2.0 leaderboard
&
Clean-to-clean / clean-to-randomized
\\

ABot-M0
&
Official RoboTwin 2.0 leaderboard
&
Clean-to-clean / clean-to-randomized
\\

X-VLA
&
Official RoboTwin 2.0 leaderboard
&
Clean-to-clean / clean-to-randomized
\\

X-WAM
&
Official RoboTwin 2.0 leaderboard
&
Clean-to-clean / clean-to-randomized
\\

$\pi_{0.5}$
&
Official RoboTwin 2.0 leaderboard
&
Clean-to-clean / clean-to-randomized
\\

\midrule

StarVLA
&
Official RoboTwin 2.0 leaderboard
&
Leaderboard entry; not the StarVLA-OFT row used in Table~\ref{tab:libero}
\\

AHA-WAM
&
Official RoboTwin 2.0 leaderboard
&
Clean-to-clean / clean-to-randomized
\\

Fast-WAM$^\dagger$
&
Our reproduction
&
See Appendix~\ref{app:reproductions}
\\

Evo-1
&
Official Evo-1 RoboTwin-50 clean-only evaluation
&
Released clean-only multi-task checkpoint
\\

PINT
&
Our evaluation
&
--
\\

PINT-VA
&
Our evaluation
&
--
\\

\bottomrule
\end{tabularx}
\end{table}

The official RoboTwin 2.0 leaderboard reports Easy and Hard as clean-to-clean and clean-to-randomized evaluation, respectively.
Table~\ref{tab:robotwin} rounds leaderboard values to the displayed precision.
Evo-1 is taken from the released clean-only RoboTwin-50 evaluation, where a single multi-task policy is evaluated for 100 rollouts per task under both settings.

\subsection{Our Reproductions and Evaluations}
\label{app:reproductions}

The Fast-WAM result marked $\dagger$ in Table~\ref{tab:robotwin} is our reproduction on RoboTwin 2.0.
We train the baseline on the same 50-task clean demonstration set used for PINT and evaluate it under the same Easy and Hard protocol.
The official RoboTwin 2.0 leaderboard contains a separate Fast-WAM entry, so we explicitly distinguish our reproduced result with $\dagger$.

The real-world $\pi_{0.5}$ and Motus results in Table~\ref{tab:real_world} are also obtained in our evaluation.
For both baselines, we initialize from their official pretrained checkpoints and fine-tune on the same 700 real-world demonstrations used to train PINT.
All three methods are then evaluated under the same ID and OOD protocol described in Appendix~\ref{app:realworld}.
In contrast, the Motus results reported in Table~\ref{tab:libero} are external public evaluations from MoWAM~\citep{wang2026mowam} and are not our reproduction.

\subsection{Future-Video Baseline Provenance}
\label{app:future_video_provenance}

The entry labeled \texttt{WAM (Baseline)} in Fig.~\ref{fig:qualitative} uses the VGM from our Fast-WAM reproduction reported in Table~\ref{tab:robotwin}.
It is trained on the same RoboTwin 2.0 clean demonstrations used for PINT and evaluated on the same ID and OOD inputs shown in the figure.
The figure uses \texttt{WAM (Baseline)} to denote the baseline WAM configuration, while the caption identifies its concrete implementation as Fast-WAM.

\FloatBarrier

\section{Additional Quantitative Results}
\label{app:additional_results}

\subsection{LIBERO and LIBERO-Plus Ablations and OOD Analysis}
\label{app:libero_ablation}

We further evaluate two key design choices across the seven distribution shifts in LIBERO-Plus.
Prompt-conditioned VGM replaces Predictive Intent with the VGM's original prompt conditioning,
while w/o Stage~1 Video Loss removes future-video supervision during Stage~1.
Table~\ref{tab:libero_ablation} reports both the standard LIBERO result and the category-level LIBERO-Plus results.

\begin{table}[!htbp]
\centering
\footnotesize
\setlength{\tabcolsep}{3.2pt}
\caption{\textbf{Ablations on LIBERO and LIBERO-Plus.}
Success rates (\%) on LIBERO (ID) and across the seven LIBERO-Plus distribution shifts.
The LIBERO-Plus Total is computed over all 10,030 evaluation trials rather than by averaging the seven category-level success rates.}
\label{tab:libero_ablation}
\begin{tabular}{@{}lccccccccc@{}}
\toprule
& LIBERO
& \multicolumn{8}{c}{LIBERO-Plus} \\
\cmidrule(lr){2-2}
\cmidrule(lr){3-10}
Setting
& Avg.
& Cam.
& Robot
& Lang.
& Light
& BG
& Noise
& Layout
& Total \\
\midrule
Prompt-conditioned VGM
& 95.9
& 53.2
& 39.1
& 78.7
& 88.3
& 45.7
& 69.8
& 70.9
& 63.4 \\

w/o Stage~1 Video Loss
& 97.1
& 53.5
& 65.3
& 80.2
& 88.7
& 88.8
& 83.1
& 75.1
& 75.2 \\

\rowcolor{gray!15}
PINT
& \textbf{98.0}
& \textbf{59.2}
& \textbf{73.7}
& \textbf{84.4}
& \textbf{94.0}
& \textbf{91.4}
& 80.8
& \textbf{81.2}
& \textbf{79.5} \\
\bottomrule
\end{tabular}
\end{table}

Replacing Predictive Intent with prompt conditioning reduces success by 2.1 points on LIBERO but by 16.1 points on LIBERO-Plus.
At the category level, the largest degradations occur under background-texture and robot-initial-state shifts, where success decreases by 45.7 and 34.6 points relative to PINT, respectively.
Removing Stage~1 future-video supervision has a substantially milder effect, reducing the overall LIBERO-Plus score by 4.3 points, with the largest drops under robot initial states (8.4 points), object layout (6.1 points), camera viewpoint (5.7 points), and lighting conditions (5.3 points).
Overall, these results suggest that Predictive Intent conditioning is particularly important
for robustness to substantial scene and initialization shifts, while Stage~1 future-video supervision
provides additional OOD robustness overall.

\subsection{Repeated Evaluation and Uncertainty}
\label{app:repeated_evaluation}

To quantify evaluation variance, we evaluate each benchmark checkpoint three times using different evaluation seeds.
For RoboTwin 2.0, each evaluation uses 100 rollouts per task over all 50 tasks.
Table~\ref{tab:repeated_evaluation} reports the individual evaluations together with their mean and standard deviation.

\begin{table}[!htbp]
\centering
\small
\caption{\textbf{Repeated evaluation of PINT.}
Each row uses a fixed trained checkpoint and three evaluation seeds.
We report the mean and sample standard deviation across evaluations.}
\label{tab:repeated_evaluation}
\begin{tabular}{llcccc}
\toprule
Benchmark & Metric & Eval.\ 1 & Eval.\ 2 & Eval.\ 3 & Mean $\pm$ Std. \\
\midrule
RoboTwin 2.0 & Easy  & 68.34 & 68.66 & 68.90 & 68.63 $\pm$ 0.28 \\
RoboTwin 2.0 & Hard  & 15.08 & 15.52 & 15.86 & 15.49 $\pm$ 0.39 \\
LIBERO        & Avg.  & 98.2  & 97.8  & 98.0  & 98.00 $\pm$ 0.20 \\
LIBERO-Plus   & Total & 79.5  & 79.5  & 79.4  & 79.47 $\pm$ 0.06 \\
\bottomrule
\end{tabular}
\end{table}

The repeated evaluations show limited variability for a fixed trained checkpoint across all three benchmarks.
In particular, the standard deviation is 0.39 points on RoboTwin 2.0 Hard and 0.06 points on LIBERO-Plus, indicating that the reported aggregate performance is stable across evaluation seeds.

\subsection{Inference Efficiency}
\label{app:inference_efficiency}

We measure inference latency for Fast-WAM, PINT, and PINT-VA under the same hardware and inference configuration.
All measurements are performed on a single NVIDIA A100 80GB GPU using BF16 precision and batch size 1.
For each model, we first run 100 inputs for warm-up, followed by three independent timing rounds of 100 inputs each.
We report the mean end-to-end latency over the resulting 300 timed runs.
The measured latency covers generation of a complete 32-step action chunk with 10 action denoising steps, including the VLM forward pass for PINT and PINT-VA.

Default PINT performs a single VGM forward pass on the current frame conditioned on Predictive Intent and caches the resulting layer-wise K/V features for all AE denoising steps.
PINT-VA additionally lets the AE access VGM features from all video latents and synchronously denoises future video and action latents.

\begin{table}[!htbp]
\centering
\small
\caption{\textbf{Performance and inference efficiency.}
RoboTwin 2.0 success rates (\%) and mean end-to-end latency for generating one 32-step action chunk.
All latency measurements use the same NVIDIA A100 80GB GPU, BF16 precision, batch size 1, and 10 action denoising steps.}
\label{tab:inference_efficiency}
\begin{tabular}{lcccc}
\toprule
Method & Easy $\uparrow$ & Hard $\uparrow$ & Avg. $\uparrow$ & Latency (ms) $\downarrow$ \\
\midrule
Fast-WAM & 72.4 & 1.8  & 37.1 & 266 \\
PINT     & 68.6 & 15.5 & 42.1 & 331 \\
PINT-VA  & 71.9 & 18.1 & 45.0 & 776 \\
\bottomrule
\end{tabular}
\end{table}

PINT increases end-to-end latency from 266 to 331 ms relative to Fast-WAM while substantially improving Hard and average success.
PINT-VA further improves manipulation performance, with higher inference cost due to synchronous future-video and action denoising.

\subsection{Per-Task Results on RoboTwin 2.0}
\label{app:robotwin_per_task}
\label{app:robotwin_full_results}

Table~\ref{tab:robotwin_full_results} reports per-task Easy and Hard results for PINT and PINT-VA alongside publicly available results for RDT, $\pi_0$, and Evo-1 over all 50 RoboTwin 2.0 tasks.
These results complement the aggregate comparisons reported in Table~\ref{tab:robotwin}.

\begin{table}[!htbp]
\centering
\footnotesize
\renewcommand{\arraystretch}{0.90}
\setlength{\tabcolsep}{5pt}

\caption{\textbf{Per-task performance on RoboTwin 2.0.}
Success rates (\%) on all 50 tasks under the Easy and Hard settings. PINT results are averaged over three repeated evaluations.}
\label{tab:robotwin_full_results}

\begin{tabular}{@{}lcccccccccc@{}}
\toprule
& \multicolumn{2}{c}{RDT}
& \multicolumn{2}{c}{$\pi_0$}
& \multicolumn{2}{c}{Evo-1}
& \multicolumn{2}{c}{\textbf{PINT}}
& \multicolumn{2}{c}{\textbf{PINT-VA}} \\
\cmidrule(lr){2-3}
\cmidrule(lr){4-5}
\cmidrule(lr){6-7}
\cmidrule(lr){8-9}
\cmidrule(lr){10-11}
Task
& Easy & Hard
& Easy & Hard
& Easy & Hard
& Easy & Hard
& Easy & Hard \\
\midrule
Adjust Bottle              & 81 & 75 & 90 & 56 & 100 & 40 & 99.67 & 40.67 & 99 & 50 \\
Beat Block Hammer          & 77 & 37 & 43 & 21 & 82  & 8 & 87.67 & 13.67 & 95 & 19 \\
Blocks Ranking RGB         & 3  & 0  & 19 & 5  & 76  & 21 & 82.67 & 3.67 & 83 & 6 \\
Blocks Ranking Size        & 0  & 0  & 7  & 1  & 58  & 7 & 34.00 & 0.00 & 53 & 2 \\
Click Alarmclock           & 61 & 12 & 63 & 11 & 98  & 18 & 84.67 & 22.33 & 93 & 17 \\
Click Bell                 & 80 & 9  & 44 & 3  & 95  & 36 & 86.67 & 22.33 & 97 & 6 \\
Dump Bin Bigbin            & 64 & 32 & 83 & 24 & 95  & 46 & 90.67 & 61.67 & 95 & 63 \\
Grab Roller                & 74 & 43 & 96 & 80 & 98  & 17 & 99.67 & 51.67 & 100 & 48 \\
Handover Block             & 45 & 14 & 45 & 8  & 74  & 3 & 56.67 & 0.00 & 78 & 1 \\
Handover Mic               & 90 & 31 & 98 & 13 & 100 & 2 & 90.67 & 3.67 & 94 & 9 \\
Hanging Mug                & 23 & 16 & 11 & 3  & 9   & 1 & 20.00 & 3.67 & 19 & 0 \\
Lift Pot                   & 72 & 9  & 84 & 36 & 81  & 7 & 85.67 & 0.33 & 92 & 0 \\
Move Can Pot               & 25 & 12 & 58 & 21 & 69  & 3 & 71.67 & 0.33 & 70 & 3 \\
Move Pillbottle Pad        & 8  & 0  & 21 & 1  & 75  & 0 & 70.67 & 0.33 & 75 & 3 \\
Move Playingcard Away      & 43 & 11 & 53 & 22 & 81  & 1 & 90.67 & 10.67 & 89 & 5 \\
Move Stapler Pad           & 2  & 0  & 0  & 2  & 0   & 0 & 25.00 & 0.00 & 18 & 0 \\
Open Laptop                & 59 & 32 & 85 & 46 & 82  & 2 & 89.67 & 7.67 & 85 & 14 \\
Open Microwave             & 37 & 20 & 80 & 50 & 70  & 4 & 72.67 & 47.33 & 67 & 20 \\
Pick Diverse Bottles       & 2  & 0  & 27 & 6  & 49  & 26 & 57.00 & 13.67 & 64 & 15 \\
Pick Dual Bottles          & 42 & 13 & 57 & 12 & 76  & 40 & 67.67 & 13.67 & 86 & 14 \\
Place A2B Left             & 3  & 1  & 31 & 1  & 48  & 5 & 46.67 & 7.67 & 46 & 14 \\
Place A2B Right            & 1  & 1  & 27 & 6  & 38  & 3 & 47.67 & 3.67 & 44 & 11 \\
Place Bread Basket         & 10 & 2  & 17 & 4  & 63  & 19 & 78.67 & 13.67 & 86 & 24 \\
Place Bread Skillet        & 5  & 1  & 23 & 1  & 63  & 9 & 86.67 & 7.67 & 84 & 7 \\
Place Burger Fries         & 50 & 27 & 80 & 4  & 90  & 23 & 96.67 & 26.33 & 98 & 40 \\
Place Can Basket           & 19 & 6  & 41 & 5  & 50  & 4 & 57.67 & 7.67 & 56 & 9 \\
Place Cans Plasticbox      & 6  & 5  & 34 & 2  & 70  & 13 & 97.67 & 13.67 & 95 & 32 \\
Place Container Plate      & 78 & 17 & 88 & 45 & 96  & 7 & 97.67 & 47.33 & 99 & 52 \\
Place Dual Shoes           & 4  & 4  & 15 & 0  & 2   & 0 & 46.67 & 19.33 & 43 & 26 \\
Place Empty Cup            & 56 & 7  & 37 & 11 & 77  & 17 & 83.67 & 7.67 & 96 & 8 \\
Place Fan                  & 12 & 2  & 20 & 10 & 34  & 0 & 48.67 & 0.33 & 47 & 0 \\
Place Mouse Pad            & 1  & 0  & 7  & 1  & 17  & 0 & 33.00 & 3.67 & 32 & 0 \\
Place Object Basket        & 33 & 17 & 16 & 2  & 79  & 6 & 57.67 & 19.00 & 69 & 20 \\
Place Object Scale         & 1  & 0  & 10 & 0  & 49  & 1 & 44.67 & 0.33 & 51 & 6 \\
Place Object Stand         & 15 & 5  & 36 & 11 & 72  & 11 & 73.67 & 30.33 & 82 & 33 \\
Place Phone Stand          & 15 & 6  & 35 & 7  & 73  & 3 & 76.67 & 0.33 & 79 & 8 \\
Place Shoe                 & 35 & 7  & 28 & 6  & 33  & 4 & 75.67 & 22.00 & 72 & 29 \\
Press Stapler              & 41 & 24 & 62 & 29 & 83  & 42 & 87.67 & 19.00 & 76 & 14 \\
Put Bottles Dustbin        & 21 & 4  & 54 & 13 & 69  & 45 & 24.00 & 0.00 & 34 & 8 \\
Put Object Cabinet         & 33 & 18 & 68 & 18 & 39  & 10 & 52.67 & 3.67 & 48 & 5 \\
Rotate QRcode              & 50 & 5  & 68 & 15 & 32  & 1 & 76.67 & 0.33 & 77 & 2 \\
Scan Object                & 4  & 1  & 18 & 1  & 32  & 0 & 62.67 & 19.00 & 68 & 27 \\
Shake Bottle Horizontally  & 84 & 51 & 99 & 51 & 98  & 28 & 97.67 & 87.67 & 97 & 81 \\
Shake Bottle               & 74 & 45 & 97 & 60 & 91  & 27 & 97.67 & 67.67 & 98 & 83 \\
Stack Blocks Three         & 2  & 0  & 17 & 0  & 87  & 3 & 40.67 & 0.33 & 65 & 1 \\
Stack Blocks Two           & 21 & 2  & 42 & 1  & 100 & 6 & 86.67 & 3.67 & 93 & 17 \\
Stack Bowls Three          & 51 & 17 & 66 & 24 & 85  & 14 & 54.67 & 3.67 & 58 & 15 \\
Stack Bowls Two            & 76 & 30 & 91 & 41 & 93  & 30 & 79.67 & 13.67 & 93 & 30 \\
Stamp Seal                 & 1  & 0  & 3  & 4  & 28  & 2 & 30.00 & 7.67 & 38 & 1 \\
Turn Switch                & 35 & 15 & 27 & 23 & 28  & 8 & 27.67 & 0.00 & 20 & 7 \\
\midrule
Average
& 34.50 & 13.72
& 46.42 & 16.34
& 65.7  & 12.5
& 68.63 & 15.49
& 71.92 & 18.10 \\
\bottomrule
\end{tabular}
\end{table}

\FloatBarrier
\section{Additional Analyses and Qualitative Results}
\label{app:additional_analysis}
\label{app:qualitative_results}

\subsection{Additional Future-Video Predictions}
\label{app:future_video}
\label{app:additional_video_results}

\begin{figure}[!htp]
    \centering
    \def\closedloopwidth{0.82\linewidth}
    \includegraphics[
        width=\closedloopwidth,
        trim={62bp 240bp 95bp 10bp},
        clip
    ]{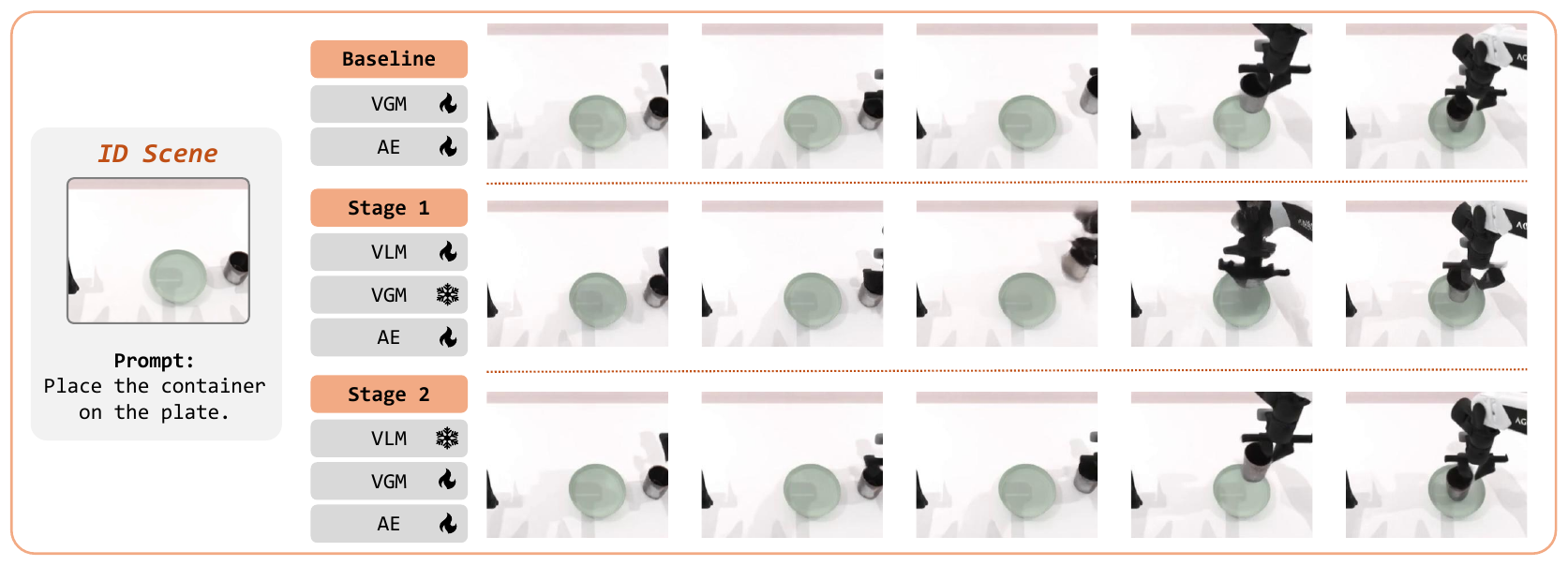}\par
    \vspace{2pt}
    {\small (a) ID closed-loop future-video predictions\par}

    \vspace{8pt}
    \includegraphics[
        width=\closedloopwidth,
        trim={62bp 240bp 95bp 10bp},
        clip
    ]{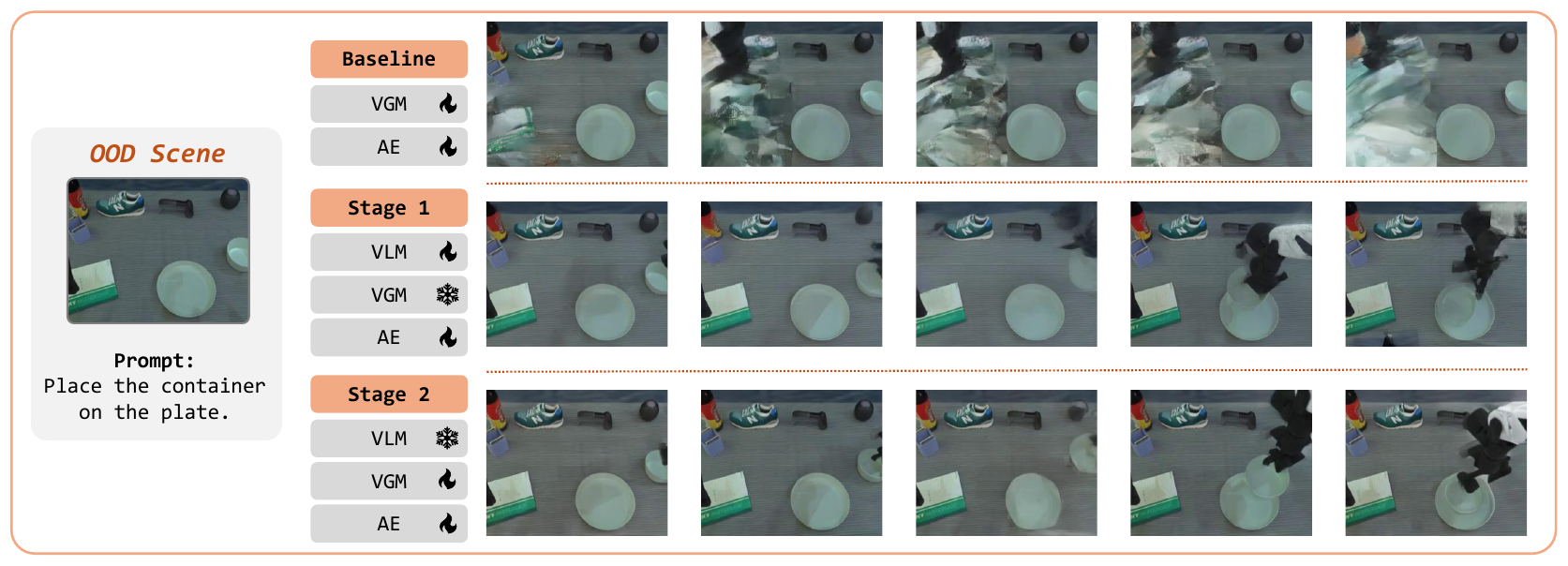}\par
    \vspace{2pt}
    {\small (b) OOD closed-loop future-video predictions\par}

    \vspace{0pt}
\caption{
\textbf{Expanded closed-loop visualization corresponding to Fig.~\ref{fig:qualitative}.}
We show closed-loop future-video predictions under ID (Easy / clean) and OOD (Hard / randomized) conditions for placing the container on the plate.
Rows compare the VGM from our Fast-WAM reproduction with PINT after Stage~1 and Stage~2.
}   
    \label{fig:app_video_closed_loop}
    \vspace{0pt}
\end{figure}

We provide two complementary qualitative views of future-video prediction.
First, Fig.~\ref{fig:app_video_closed_loop} expands the example in Fig.~\ref{fig:qualitative} and shows the corresponding closed-loop future-video predictions under ID and OOD conditions for the task of placing the container on the plate.
This closed-loop view shows how the predictions evolve during policy execution for the VGM trained in Fast-WAM and PINT after Stage~1 and Stage~2.

\suppressfloats[t]

\begin{figure}[!htp]
    \centering
    \includegraphics[
        width=0.66\linewidth,
        trim={85bp 170bp 110bp 35bp},
        clip
    ]{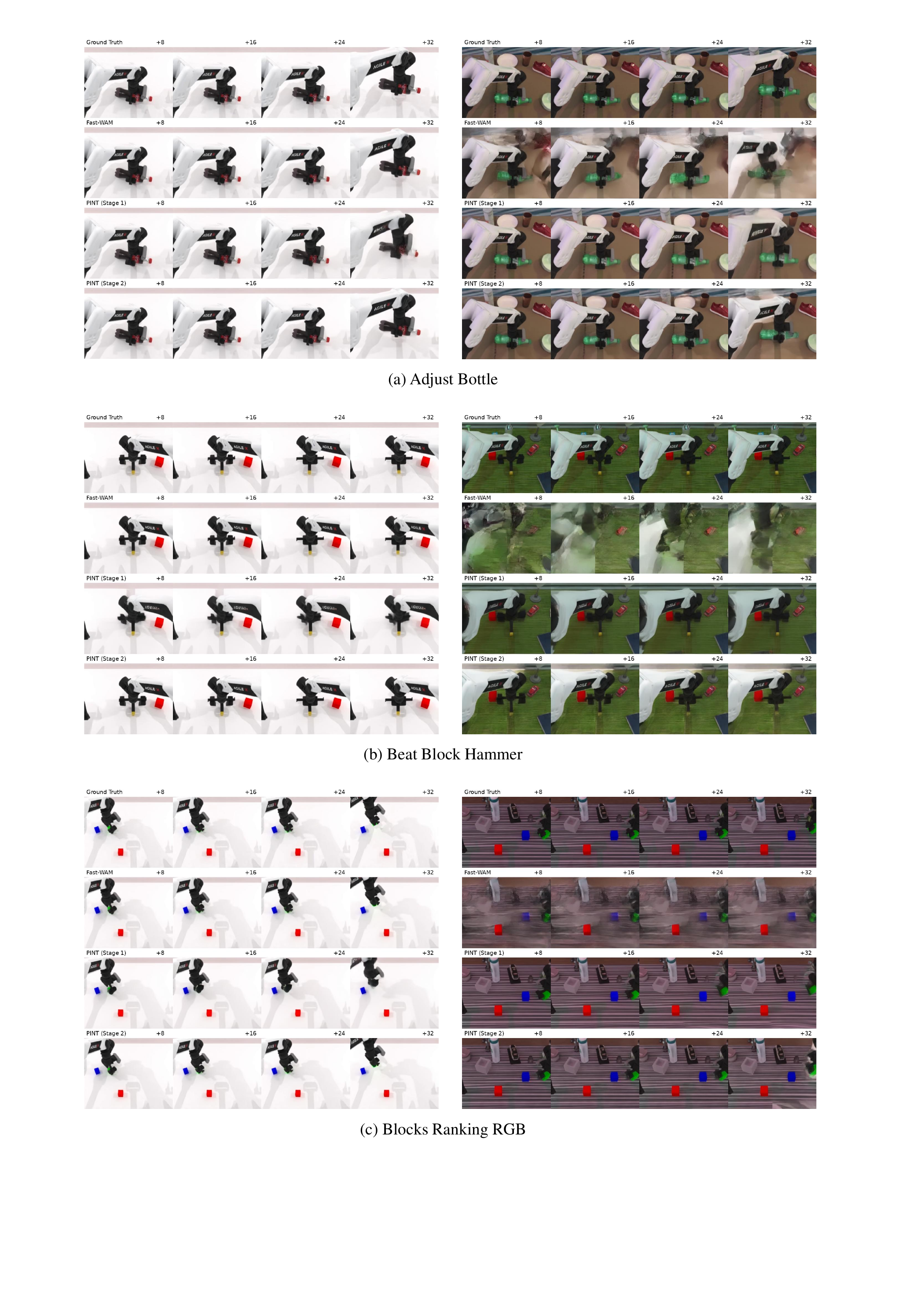}\par
    \vspace{0pt}
\caption{
\textbf{Additional open-loop future-video predictions on RoboTwin 2.0 (part I).}
For each task, the left and right panels show ID (Easy / clean) and OOD (Hard / randomized) examples, respectively.
Rows show the ground-truth future frames, the VGM trained in Fast-WAM, PINT after Stage~1, and PINT after Stage~2.
Columns show the main-camera view at $t{+}8$, $t{+}16$, $t{+}24$, and $t{+}32$.
}
    \label{fig:app_video_a}
    \vspace{0pt}
\end{figure}

\begin{figure}[!htp]
    \centering
    \includegraphics[
        width=0.66\linewidth,
        trim={85bp 580bp 110bp 35bp},
        clip
    ]{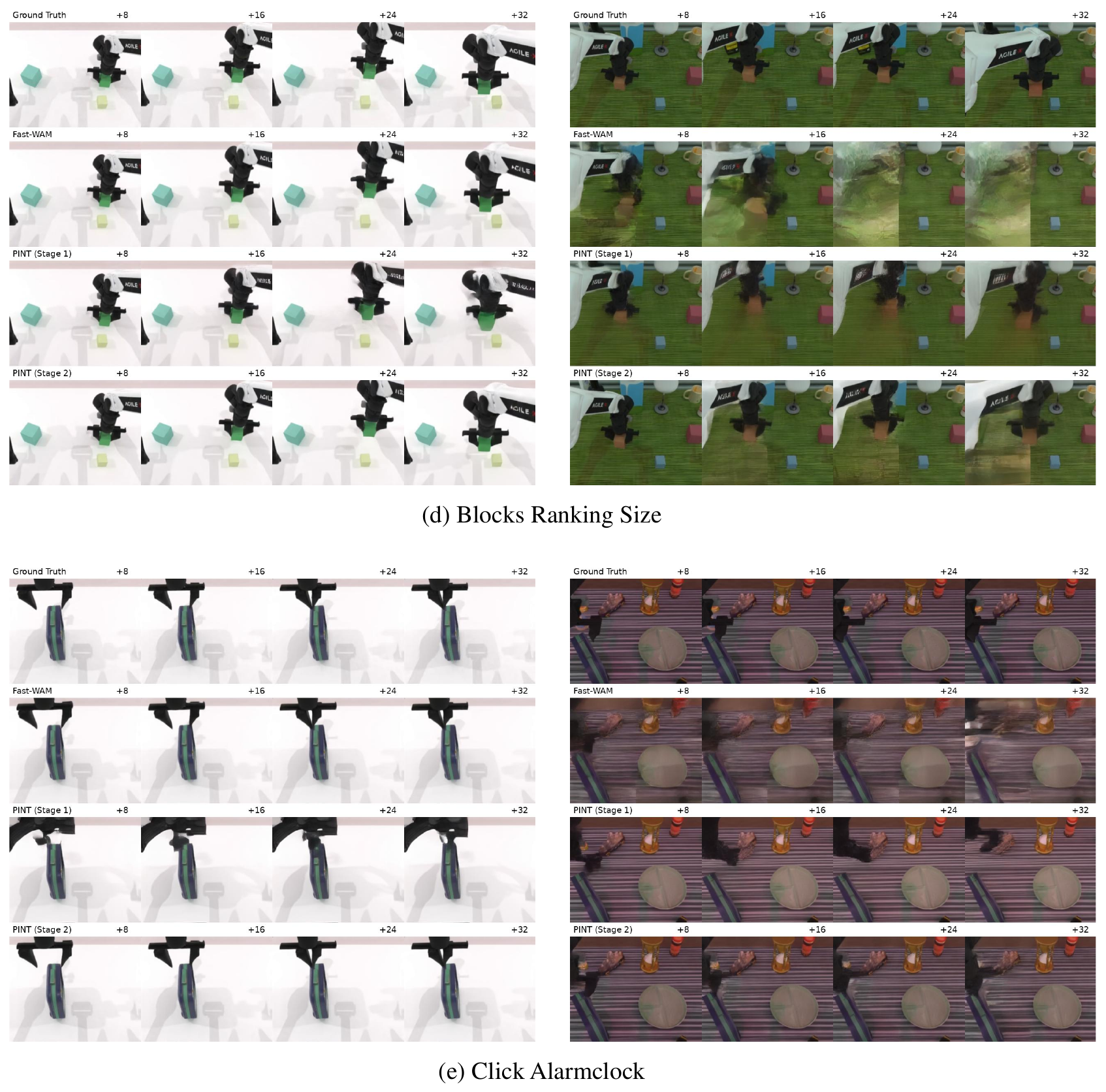}\par
    \vspace{0pt}
    \caption{
    \textbf{Additional open-loop future-video predictions on RoboTwin 2.0 (part II).}
    Same visualization format as Fig.~\ref{fig:app_video_a}.
    }
    \label{fig:app_video_b}
    \vspace{0pt}
\end{figure}

We additionally evaluate future-video prediction in a controlled open-loop setting using expert demonstrations.
Figs.~\ref{fig:app_video_a} and~\ref{fig:app_video_b} show the first five tasks in the fixed RoboTwin 2.0 task list.
For each task and evaluation setting, we use a fixed demonstration episode and choose the prediction start time at the midpoint of its valid prediction range.
All models receive the same current frame and task instruction and predict future frames at $t{+}8$, $t{+}16$, $t{+}24$, and $t{+}32$, which are compared with the corresponding ground-truth frames from the same demonstration.
The ID and OOD examples are drawn from clean and randomized demonstrations, respectively, and therefore need not share the same scene layout.

\clearpage

\FloatBarrier

In the ID examples, the VGM trained in Fast-WAM and both PINT stages generally produce predictions close to the ground truth.
Under OOD shifts, the VGM trained in Fast-WAM often shows stronger degradation in scene appearance and structure, particularly for Beat Block Hammer and the Blocks Ranking tasks.
PINT better preserves the scene layout in these examples, while Stage~2 generally refines the predicted arm and object motion relative to Stage~1.
Prediction quality decreases at the longest horizon for all methods, with visible blurring in some $t{+}32$ frames.

\subsection{Predictive Intent Visualization Details}
\label{app:intent_visualization}

The visualization in Fig.~\ref{fig:intent_visualization} is constructed from RoboTwin 2.0 clean training demonstrations.
We extract Predictive Intent every 10 data frames and retain samples with a complete 32-step future action chunk.
For each retained sample, the visualization pairs the Predictive Intent and ground-truth action chunk from the same episode and timestep.

For the two-dimensional coordinates, we use the ground-truth continuous action chunk of shape $32\times14$.
Each action dimension is standardized across the retained samples, after which the chunk is flattened, reduced to 50 dimensions with PCA, and projected to two dimensions with t-SNE.

Colors are computed independently from Predictive Intent.
We take the final-layer hidden states of the 32 Intent Tokens before the expert-specific projections and mean-pool them across tokens.
The resulting representation is projected to three PCA dimensions, which are mapped to the RGB channels.
Each channel is independently scaled using its 1st and 99th percentiles before clipping to the valid color range.

The Intent representations are extracted from the final Stage~1 checkpoint.
Because Stage~2 freezes the VLM and Intent Tokens, the Predictive Intent representation remains fixed during subsequent expert adaptation.
The highlighted regions in Fig.~\ref{fig:intent_visualization} are used only for qualitative inspection of the corresponding demonstration segments and do not represent supervised behavior labels or automatically discovered semantic clusters.

\clearpage
\subsection{Real-World Qualitative Rollouts}
\label{app:realworld_qualitative}

We provide qualitative real-world rollouts to complement the quantitative evaluation of PINT.
The visualizations illustrate representative executions of the five manipulation tasks under both ID and OOD conditions.
For each rollout, temporally ordered observations are shown from left to right to illustrate task progression.

\begin{figure}[!ht]
\centering
\includegraphics[
width=0.58\linewidth,
trim={40pt 36pt 46pt 14pt},
clip
]{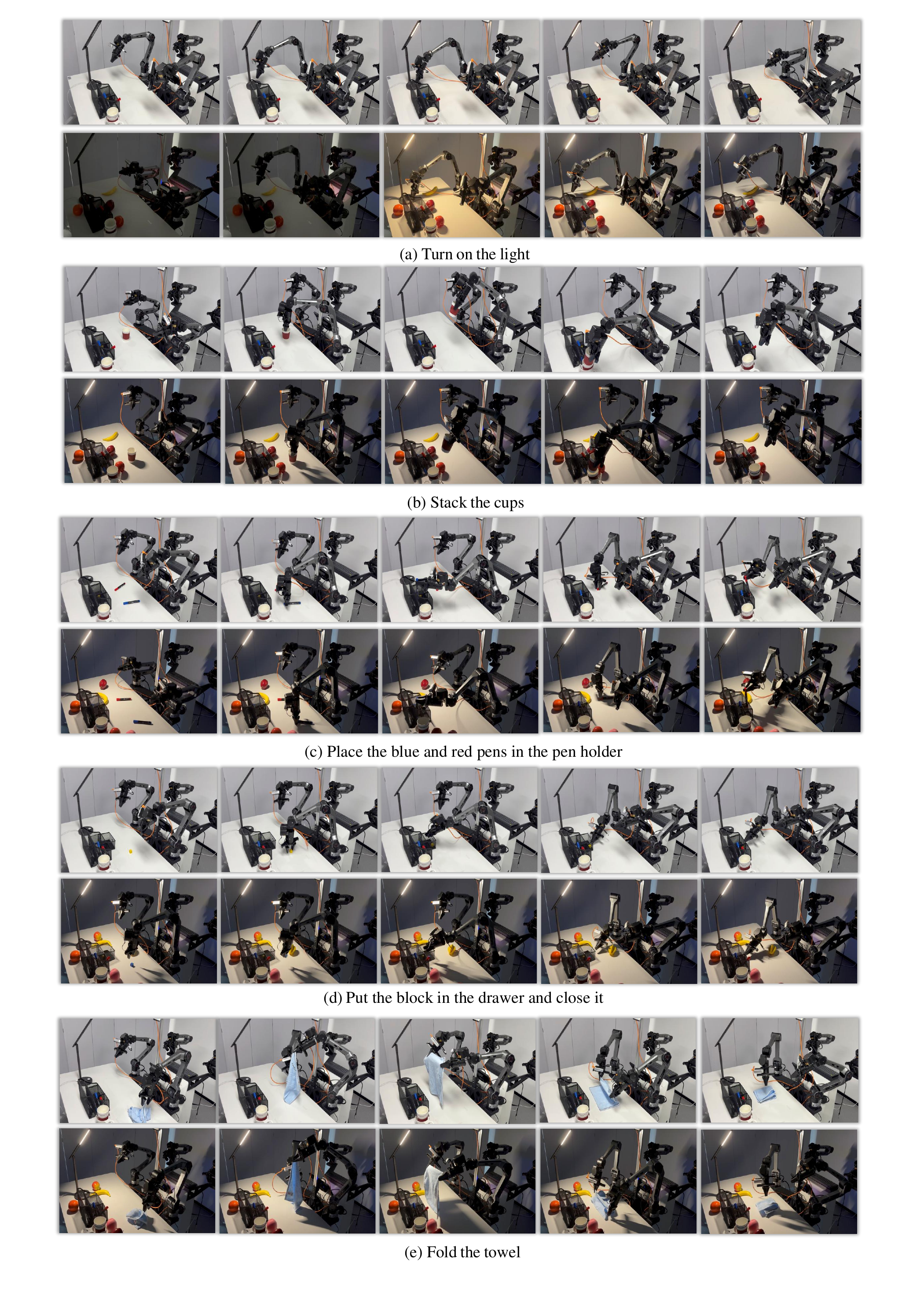}\par
\vspace{2pt}
\caption{\textbf{Qualitative real-world rollouts on individual tasks.}
We show representative executions of Turn On Light, Stack Cups, Place Pens, Store Block, and Fold Towel under both in-distribution and out-of-distribution conditions.
For each task, the top row shows the in-distribution rollout and the bottom row shows the corresponding out-of-distribution rollout.
Frames are ordered temporally from left to right.
}
\label{fig:realworld_subtasks}
\end{figure}

\end{document}